\documentclass[11pt]{article}

\usepackage[preprint]{acl}

\usepackage{times}
\usepackage{latexsym}

\usepackage[T1]{fontenc}

\usepackage[utf8]{inputenc}

\usepackage{microtype}

\usepackage{inconsolata}

\usepackage{graphicx}
\usepackage{makecell}
\usepackage{rotating}
\usepackage{enumitem}
\usepackage{bbm}

\usepackage{caption}
\usepackage{subcaption}

\usepackage{booktabs}
\usepackage{multirow}
\usepackage{amsmath}
\usepackage{amsfonts}
\usepackage[table]{xcolor}
\usepackage{color,colortbl}
\usepackage{tabularx}
\usepackage{algorithm}
\usepackage{algpseudocode}
\usepackage{amsmath}

\usepackage{pgfplots}
\pgfplotsset{compat=1.18}

\usepackage{mdframed}
\usepackage{lipsum}
\usepackage{tcolorbox}
\tcbuselibrary{theorems}
\usepackage{ulem} 

\definecolor{darkyellow}{RGB}{180,140,0}
\definecolor{gold}{RGB}{212,175,55}
\definecolor{tablecolor}{RGB}{251,198,194}

\definecolor{color_understanding}{HTML}{add8e6}
\definecolor{color_process}{HTML}{E59693}
\definecolor{color_check}{HTML}{6CC3B5}
\definecolor{color_answer}{HTML}{dfbf7f}

\title{Hidden Language Consistency Phenomena in Reasoning LLMs}

\author{Muhammad Ali Shafique \\
  Kansas State University \\
  \texttt{alishafique@ksu.edu} \\\And
  Kelly Marchisio \\
  Cohere \\
  \texttt{kelly@cohere.com} \\}

\begin{document}
\maketitle

\begin{abstract}
Multilingual reasoning models are commonly evaluated by whether they arrive at the correct answer, but not by whether they preserve the intended language while reasoning and responding. This omission conceals important multilingual behaviors that emerge as tasks become harder. In this paper, we study task difficulty, task accuracy, thinking-language consistency (TC), and answer-language consistency (AC) across reasoning models using PolyMath benchmark in eight languages and four difficulty levels. We uncover four findings: (1) language consistency exhibits four difficulty-dependent behaviors: output-language consistency remains aligned with input, remains misaligned, degrades gradually, or collapses abruptly. (2) We identify the \textit{language consistency breakdown effect}, where increasing difficulty can cause a sudden drop in output-language consistency, especially in less strongly represented and non-Latin-script languages. (3) Due to this breakdown effect, accuracy can be preserved or even improved at a harder difficulty level as the model shifts to its internal dominant language. (4) Quantization can improve or degrade output-language consistency independently of its effect on accuracy, with GPTQ and AWQ often outperforming AutoRound under tolerance-based voting with $\epsilon=1.0$. These results show that multilingual capability cannot be characterized by accuracy alone; reliable evaluation should jointly consider task accuracy, language consistency, and task difficulty for multilingual benchmarks.
\end{abstract}






\section{Introduction}
\label{sec:intro}

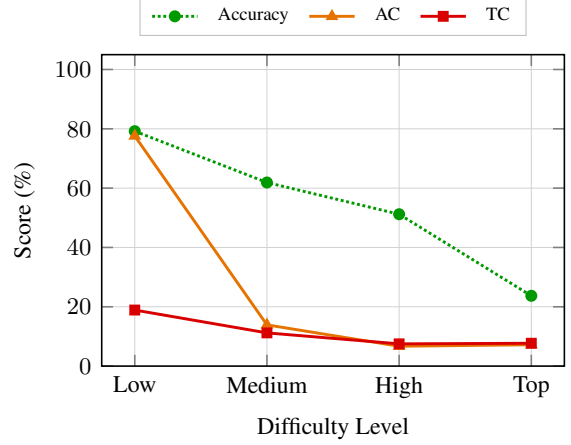
\begin{figure}[t]
\centering
\begin{tikzpicture}
\begin{axis}[
    width=\columnwidth,
    height=5.7cm,
    xlabel={Difficulty Level},
    ylabel={Score (\%)},
    xmin=0.75, xmax=4.25,
    ymin=0, ymax=105,
    xtick={1,2,3,4},
    xticklabels={Low, Medium, High, Top},
    ytick={0,20,40,60,80,100},
    tick label style={font=\small},
    label style={font=\small},
    axis line style={line width=0.6pt},
    tick style={line width=0.6pt},
    grid=major,
    major grid style={
        line width=0.25pt,
        draw=gray!35,
    },
    legend style={
        at={(0.5,1.06)},
        anchor=south,
        legend columns=3,
        draw=black!25,
        fill=white,
        font=\scriptsize,
        column sep=0.18cm,
        inner xsep=4pt,
        inner ysep=2pt,
    },
]

\addplot[
    color=green!60!black,
    densely dotted,
    mark=*,
    mark options={solid},
    line width=1.1pt,
    mark size=1.7pt
] coordinates {
    (1,79.2) (2,61.9) (3,51.2) (4,23.7)
};
\addlegendentry{Accuracy}

\addplot[
    color=orange!90!black,
    solid,
    mark=triangle*,
    mark options={solid},
    line width=1.1pt,
    mark size=1.7pt
] coordinates {
    (1,77.6) (2,13.9) (3,6.7) (4,7.2)
};
\addlegendentry{AC}

\addplot[
    color=red!85!black,
    solid,
    mark=square*,
    mark options={solid},
    line width=1.1pt,
    mark size=1.5pt
] coordinates {
    (1,18.9) (2,11.2) (3,7.5) (4,7.7)
};
\addlegendentry{TC}

\end{axis}
\end{tikzpicture}

\caption{
Difficulty-wise accuracy, thinking-language consistency (TC), and answer-language consistency (AC) of OLMo-3-7B-Think on Chinese PolyMath problems. While accuracy decreases progressively with task difficulty, AC exhibits a sharp breakdown from 77.6\% to 13.9\% between the low and medium levels. TC remains consistently low across all difficulty levels.
}
\label{fig:olmo_zh_acc_tc_ac_difficulty_teaser}
\end{figure}

Recent large language models (LLMs) with enhanced reasoning capabilities have achieved strong performance on mathematical, scientific, and competition-level reasoning tasks~\citep{jaech2024openai,team2025kimi,guo2025deepseek,qwq32b,rein2024gpqa,lightman2023let,cobbe2021gsm8k}. These gains often rely on long chain-of-thought reasoning~\citep{wei2022chain}, which can improve accuracy but also increase inference cost and lead models to ``overthink'' simple problems~\citep{chen2024not}. As reasoning models are increasingly used across languages, evaluation must consider not only whether the final answer is correct, but also whether the model preserves the requested language in its reasoning and/or response.

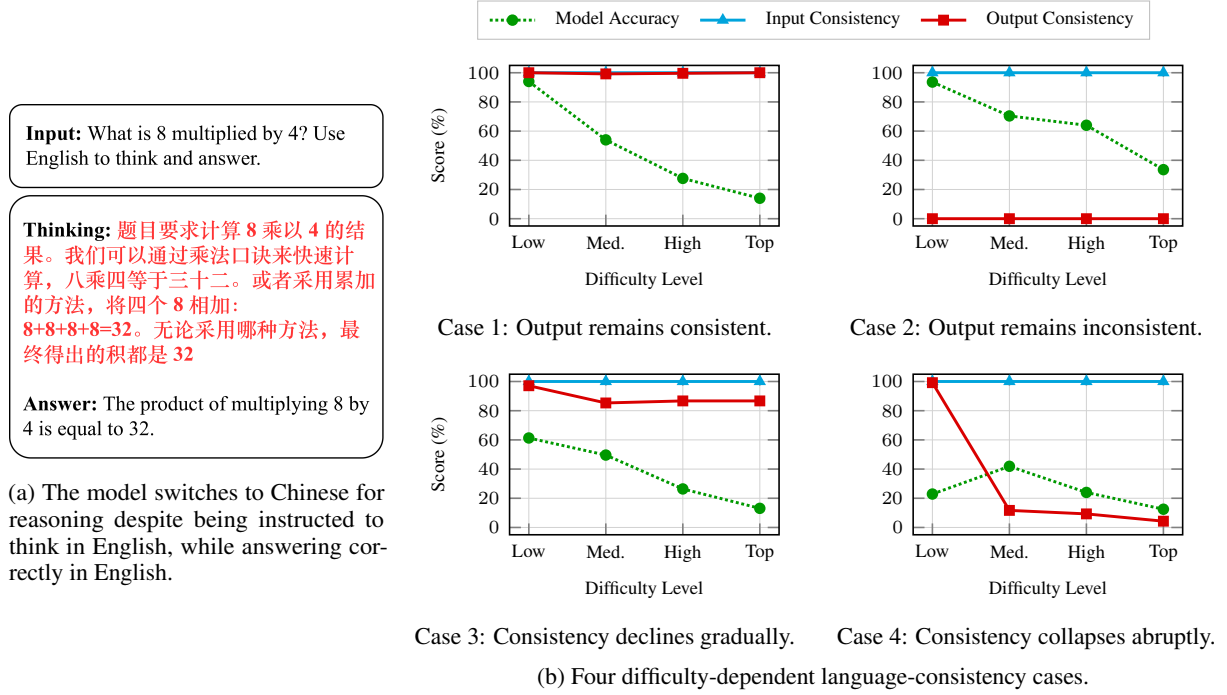
\begin{figure*}[t]
\centering

\textbf{Four Input--Output Language-Consistency Cases across Task Difficulty Levels}

\vspace{0.50cm}

\pgfplotsset{
    languagephenomenon/.style={
        width=\linewidth,
        height=0.72\linewidth,
        xlabel={Difficulty Level},
        xmin=0.75, xmax=4.25,
        ymin=-5, ymax=105,
        xtick={1,2,3,4},
        xticklabels={Low, Med., High, Top},
        ytick={0,20,40,60,80,100},
        tick label style={font=\scriptsize},
        label style={font=\scriptsize},
        axis line style={line width=0.6pt},
        tick style={line width=0.6pt},
        grid=major,
        major grid style={
            line width=0.25pt,
            draw=gray!35
        },
    }
}

\begin{subfigure}[c]{0.31\textwidth}
\centering

\begin{minipage}[c][0.54\textwidth][c]{\linewidth}
\centering
\includegraphics[width=\linewidth]{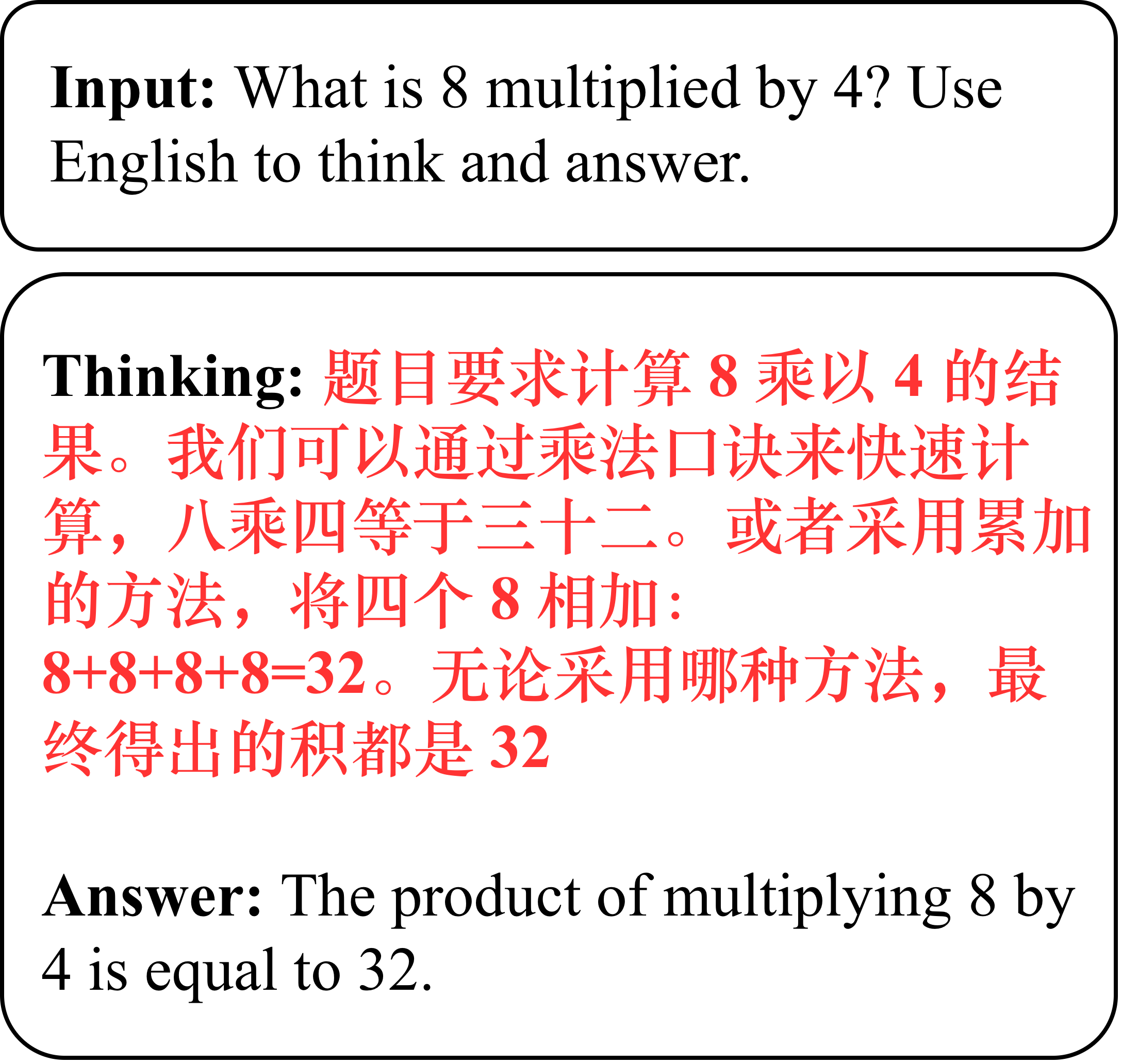}

\vspace{0.08cm}

\caption{The model switches to Chinese for reasoning despite being instructed to think in English, while answering correctly in English.}
\label{fig:language-inconsistent-example}
\end{minipage}

\end{subfigure}
\hfill
\begin{subfigure}[c]{0.67\textwidth}
\centering

\begin{tikzpicture}
\begin{axis}[
    hide axis,
    xmin=0, xmax=1,
    ymin=0, ymax=1,
    width=\linewidth,
    height=2cm,
    legend columns=3,
    legend style={
        draw=black!25,
        fill=white,
        font=\scriptsize,
        at={(0.5,0.5)},
        anchor=center,
        column sep=0.18cm,
        inner xsep=4pt,
        inner ysep=2pt
    },
]

\addlegendimage{
    color=green!60!black,
    densely dotted,
    mark=*,
    mark options={solid},
    line width=1.1pt,
    mark size=1.6pt
}
\addlegendentry{Model Accuracy}

\addlegendimage{
    color=cyan!90!black,
    solid,
    mark=triangle*,
    mark options={solid},
    line width=1.1pt,
    mark size=1.6pt
}
\addlegendentry{Input Consistency}

\addlegendimage{
    color=red!85!black,
    solid,
    mark=square*,
    mark options={solid},
    line width=1.1pt,
    mark size=1.5pt
}
\addlegendentry{Output Consistency}

\end{axis}
\end{tikzpicture}

\vspace{-0.25cm}

\begin{minipage}[t]{0.48\linewidth}
\vspace{0pt}
\centering

\begin{tikzpicture}
\begin{axis}[
    languagephenomenon,
    ylabel={Score (\%)}
]

\addplot[
    color=green!60!black,
    densely dotted,
    mark=*,
    mark options={solid},
    line width=1.1pt,
    mark size=1.6pt
] coordinates {
    (1,94.0) (2,54.0) (3,27.6) (4,14.0)
};

\addplot[
    color=cyan!90!black,
    solid,
    mark=triangle*,
    mark options={solid},
    line width=1.1pt,
    mark size=1.6pt
] coordinates {
    (1,100.0) (2,100.0) (3,100.0) (4,100.0)
};

\addplot[
    color=red!85!black,
    solid,
    mark=square*,
    mark options={solid},
    line width=1.1pt,
    mark size=1.5pt
] coordinates {
    (1,100.0) (2,99.2) (3,99.6) (4,100.0)
};

\end{axis}
\end{tikzpicture}

{\small{Case 1:} Output remains consistent.}
\end{minipage}
\hfill
\begin{minipage}[t]{0.48\linewidth}
\vspace{0pt}
\centering

\begin{tikzpicture}
\begin{axis}[languagephenomenon]

\addplot[
    color=green!60!black,
    densely dotted,
    mark=*,
    mark options={solid},
    line width=1.1pt,
    mark size=1.6pt
] coordinates {
    (1,93.6) (2,70.4) (3,64.0) (4,33.6)
};

\addplot[
    color=cyan!90!black,
    solid,
    mark=triangle*,
    mark options={solid},
    line width=1.1pt,
    mark size=1.6pt
] coordinates {
    (1,100.0) (2,100.0) (3,100.0) (4,100.0)
};

\addplot[
    color=red!85!black,
    solid,
    mark=square*,
    mark options={solid},
    line width=1.1pt,
    mark size=1.5pt
] coordinates {
    (1,0.0) (2,0.0) (3,0.0) (4,0.0)
};

\end{axis}
\end{tikzpicture}

{\small{Case 2:} Output remains inconsistent.}
\end{minipage}

\vspace{0.30cm}

\begin{minipage}[t]{0.48\linewidth}
\vspace{0pt}
\centering

\begin{tikzpicture}
\begin{axis}[
    languagephenomenon,
    ylabel={Score (\%)}
]

\addplot[
    color=green!60!black,
    densely dotted,
    mark=*,
    mark options={solid},
    line width=1.1pt,
    mark size=1.6pt
] coordinates {
    (1,61.3) (2,49.6) (3,26.4) (4,13.1)
};

\addplot[
    color=cyan!90!black,
    solid,
    mark=triangle*,
    mark options={solid},
    line width=1.1pt,
    mark size=1.6pt
] coordinates {
    (1,100.0) (2,100.0) (3,100.0) (4,100.0)
};

\addplot[
    color=red!85!black,
    solid,
    mark=square*,
    mark options={solid},
    line width=1.1pt,
    mark size=1.5pt
] coordinates {
    (1,97.1) (2,85.3) (3,86.7) (4,86.7)
};

\end{axis}
\end{tikzpicture}

{\small{Case 3:} Consistency declines gradually.}
\end{minipage}
\hfill
\begin{minipage}[t]{0.48\linewidth}
\vspace{0pt}
\centering

\begin{tikzpicture}
\begin{axis}[languagephenomenon]

\addplot[
    color=green!60!black,
    densely dotted,
    mark=*,
    mark options={solid},
    line width=1.1pt,
    mark size=1.6pt
] coordinates {
    (1,22.9) (2,41.9) (3,24.0) (4,12.5)
};

\addplot[
    color=cyan!90!black,
    solid,
    mark=triangle*,
    mark options={solid},
    line width=1.1pt,
    mark size=1.6pt
] coordinates {
    (1,100.0) (2,100.0) (3,100.0) (4,100.0)
};

\addplot[
    color=red!85!black,
    solid,
    mark=square*,
    mark options={solid},
    line width=1.1pt,
    mark size=1.5pt
] coordinates {
    (1,99.2) (2,11.7) (3,9.3) (4,4.3)
};

\end{axis}
\end{tikzpicture}

{\small{Case 4:} Consistency collapses abruptly.}
\end{minipage}

\caption{Four difficulty-dependent language-consistency cases.}
\label{fig:language-consistency-taxonomy}
\end{subfigure}

\caption{
Illustration and taxonomy of input--output language consistency in thinking traces.
(a) An example of language switching within a thinking trace.
(b) Four representative difficulty-dependent cases:
Case~1, persistent alignment for English with Phi-4-mini-reasoning;
Case~2, persistent misalignment for Arabic with Qwen3-30B-A3B-Thinking;
Case~3, gradual degradation for Russian with DeepSeek-R1-Distill-Qwen-7B; and
Case~4, abrupt collapse for Telugu with DeepSeek-R1-Distill-Qwen-7B.
}
\label{fig:cases-language-consistency-phenomena-cases}

\end{figure*}

Multilingual reasoning is commonly evaluated through accuracy across languages and difficulty levels. However, accuracy alone can conceal important multilingual behavior. A model may produce a correct answer while reasoning or responding in an unintended input language, or preserve the requested language while producing an incorrect answer. Final-answer accuracy, thinking-language consistency (TC), and answer-language consistency (AC) therefore capture distinct but complementary dimensions of multilingual reasoning.

Task difficulty may further shape this relationship. Harder problems often require longer and more complex reasoning, increasing the likelihood of language drift. Language-specific factors, including script, tokenizer coverage, writing direction, and multilingual representation, may further amplify this behavior~\citep{ogueji-etal-2022-intriguing,nllb2022,khondaker2023gptaraeval,pozzobon2024many}. PolyMath~\citep{wang2025polymath} reports language-level output consistency scores and aggregate output consistency score across difficulty levels, but does not characterize how each output-language consistency scores evolves across difficulty levels. We fill this gap with a four-case taxonomy and language-level consistency--difficulty analysis. 

As difficulty increases, output-language consistency may remain aligned with the input language, decline gradually, or collapse abruptly as models shift toward a dominant internal language (Figure~\ref{fig:olmo_zh_acc_tc_ac_difficulty_teaser}). We term this abrupt decline the \textit{language consistency breakdown effect}.

In this paper, we investigate how task difficulty affects multilingual language consistency and how this behavior interacts with task accuracy. We evaluate five reasoning models and three non-reasoning models on PolyMath~\citep{wang2025polymath}, covering eight languages and four difficulty levels. The selected languages span diverse scripts and levels of multilingual representation. We measure final-answer accuracy, thinking-language consistency (TC), and answer-language consistency (AC) at each difficulty level, identify various language consistency behaviors, \textit{language-consistency breakdown effect}, prompt control, and systematically examine how quantization changes  output-language consistency scores. We find that:

\begin{enumerate}[leftmargin=20pt]

\item \textbf{Language consistency exhibits four distinct difficulty-dependent behaviors (\S\ref{sec:results-four-behaviors}):}
We identify a taxonomy of multilingual generation behavior across task difficulty: (1) output stays consistent with input language-consistency, (2) output stays inconsistent, (3) output-language consistency declines gradually, and (4) output-language consistency collapses abruptly. This taxonomy shows that output-language consistency degradation is not a single uniform phenomenon, but can emerge differently across models and languages (Figure~\ref{fig:cases-language-consistency-phenomena-cases}).

\item \textbf{Increasing task difficulty degrades output-language consistency and can trigger abrupt drop (\S\ref{sec:results-language-consistency-breakdown}):}
As problems become harder, output-language consistency generally declines. When it exhibits a sudden and substantial drop between consecutive difficulty levels, we term this the \textit{language consistency breakdown effect}, which is more prevalent in lower-resource and non-Latin-script languages (Case 4 in Figure~\ref{fig:language-consistency-taxonomy}).

\item \textbf{Language consistency breakdown breaks the expected monotonic accuracy--difficulty relationship (\S\ref{sec:results-nonlinear-breakdown}):}
Model accuracy is generally expected to decline monotonically as task difficulty increases. We uncover non-linear behavior that breaks this rule: accuracy can preserve or improve at a higher difficulty level due to language consistency breakdown effect in thinking-language consistency (TC) and/or answer-language consistency (AC).

\item \textbf{Quantization affects output-language consistency (\S\ref{sec:analysis_quantization}):}
To the best of our knowledge, this is the first study to systematically examine how quantization affects output-language consistency in multilingual reasoning models. We find that quantization can either improve or degrade language consistency, independently of its effect on task accuracy. Under tolerance-based voting with $\epsilon=1.0$, GPTQ and AWQ often preserves or improves output-language consistency more effectively than AutoRound.



\end{enumerate}

Overall, our work highlights that multilingual models cannot be evaluated by final-answer accuracy alone; task difficulty, thinking-language consistency, and answer-language consistency should be considered jointly to better understand multilingual reasoning behavior.

\section{Preliminary and Related Work}
\label{sec:prelim_related}

\begin{table*}[t]
\centering
\small
\setlength{\tabcolsep}{5.5pt}
\begin{tabular}{llccc}
\toprule
\textbf{Model} & \textbf{Type} &
\textbf{Avg. Acc.} &
\textbf{TC below $\tau_{\text{cons}}$ @ Top} &
\textbf{AC below $\tau_{\text{cons}}$ @ Top} \\
\midrule

DeepSeek-R1-Distill-Qwen-1.5B
& Reasoning & 20.3\scriptsize{$\pm$1.1} & 6/8 & 6/8 \\

DeepSeek-R1-Distill-Qwen-7B
& Reasoning & 36.5\scriptsize{$\pm$1.5} & 6/8 & 6/8 \\

OLMo-3-7B-Think
& Reasoning & 42.7\scriptsize{$\pm$1.0} & 6/8 & 7/8 \\

Phi-4-mini-reasoning
& Reasoning & 32.5\scriptsize{$\pm$1.1} & 6/8 & 6/8 \\

Qwen3-30B-A3B-Thinking-2507
& Reasoning & 62.3\scriptsize{$\pm$0.7} & 5/8 & 0/8 \\

\midrule

Gemma-3-12b-it
& Non-reasoning & 31.3\scriptsize{$\pm$0.6} & N/A & 0/8 \\

Phi-4-mini-instruct
& Non-reasoning & 15.6\scriptsize{$\pm$1.2} & N/A & 1/8 \\

Qwen3-30B-A3B-Instruct-2507
& Non-reasoning & 50.4\scriptsize{$\pm$0.8} & N/A & 0/8 \\

\bottomrule
\end{tabular}

\caption{
Overall task accuracy and language-consistency behavior across reasoning and non-reasoning models using forced-target-language prompting. Avg. Acc. denotes average PolyMath accuracy across eight languages. TC Below $\tau_{\text{cons}}$ @ Top and AC Below $\tau_{\text{cons}}$ @ Top count the languages with thinking- and answer-language consistency scores below $\tau_{\text{cons}}=90\%$ at Top difficulty. The threshold $\tau_{\text{cons}}$ is an analysis hyperparameter.
}
\label{tab:main-table-difficulty_breakdown}
\end{table*}

\paragraph{Multilingual Reasoning.}

Multilingual reasoning requires models to solve problems correctly while preserving the requested language in the reasoning trace, final answer, or both. Prior work shows that multilingual LLM performance varies substantially across languages because of differences in training data, tokenizer coverage, script, and degree of multilingual representation~\citep{ahia2021lowresource,ogueji-etal-2022-intriguing,nllb2022,khondaker2023gptaraeval,pozzobon2024many}. These differences may become more pronounced in reasoning tasks, where longer and more complex generations create additional chances for language drift.

Recent reasoning-oriented LLMs, such as OpenAI o1~\citep{jaech2024openai}, DeepSeek-R1~\citep{guo2025deepseek}, QwQ~\citep{qwq32b, qwen2.5technicalreport}, and Kimi~\citep{team2025kimi}, have shown strong performance on mathematical and multi-step reasoning tasks. However, their evaluation commonly emphasizes final-answer accuracy, with less attention to whether the reasoning trace and final answer remain in the requested language.


\paragraph{Code-Switching.}

Code-switching refers to natural alternation between languages within an utterance or conversation~\citep{dogruoz-etal-2021-survey}. Prior work has evaluated code-switched language understanding and generation on tasks such as sentiment analysis, machine translation, summarization, and language identification~\citep{khanuja-etal-2020-gluecos,winata-etal-2023-decades}. These studies typically use human-produced data where switching between English and another language is intentional and linguistically meaningful.

Despite this progress, current models still struggle to understand and generate code-switched text for several languages~\citep{yong-etal-2023-prompting,zhang-etal-2023-multilingual}. Our focus differs from natural code-switching because we study unintended language changes in model generations. 

\paragraph{Language Confusion.}

Unintended generation in the wrong language has been studied as source-language hallucination in zero-shot cross-lingual transfer~\citep{vu-etal-2022-overcoming,li-murray-2023-zero,pfeiffer-etal-2023-mmt5,chirkova-nikoulina-2024-key} and as off-target translation in multilingual machine translation~\citep{chen-etal-2023-target,sennrich-etal-2024-mitigating}. These behaviors occur when multilingual models generate in a language different from the requested target language in input prompt.

Similar language confusion has also been observed in LLM responses~\citep{marchisio-etal-2024-understanding,kew2023turning,faisal-anastasopoulos-2023-geographic,chen-etal-2024-monolingual,holtermann2024evaluating}. Most prior work studies this behavior at the response level. In contrast, we separately evaluate language consistency in reasoning traces and final answers. This distinction is important because a model may reason in one language and answer in another.




\section{Experimental Setup and Dataset}
\label{sec:experimental_setup}

\paragraph{Evaluation Benchmark.}
We evaluate the models on PolyMath ~\citep{wang2025polymath}, a multilingual mathematical benchmark spanning broad difficulty ranges: ``low'': K-12 Mathematics, ``medium'': High-School and University Level, ``high'': competition math, and ``top": Top Olympiad and Frontier Mathematics.  We evaluate model performance across eight languages and study the effect of difficulty levels on accuracy and language consistency in reasoning and answer traces.  PolyMath is selected because it spans diverse languages and difficulty levels and provides standard scripts\footnote{The official PolyMath scripts are used to compute final-answer correctness and language-consistency scores.} for evaluating final-answer correctness and language consistency. 

\paragraph{Models.}
We evaluate five reasoning models and three non-reasoning models: DeepSeek-R1-Distill-Qwen-1.5B, DeepSeek-R1-Distill-Qwen-7B~\citep{guo2025deepseek}, OLMo-3-7B-Think~\citep{olmo2025olmo3}, Phi-4-mini-reasoning~\cite{abdin2025phi4reasoningtechnicalreport}, and Qwen3-30B-A3B-Thinking-2507~\cite{qwen3technicalreport} for reasoning, and Gemma-3-12B-it~\cite{gemmateam2025gemma3technicalreport}, Phi-4-mini-instruct, and Qwen3-30B-A3B-Instruct-2507 as non-reasoning models.
This model selection allows comparison of multilingual behavior across reasoning and non-reasoning settings, and model scales.


\paragraph{Inference Setup and Prompting Strategy.}
All inference is performed using the \texttt{vLLM}~\cite{kwon2023efficient} framework.
We use the same inference settings as \cite{liu2025quantizationhurtsreasoningempirical}, with temperature $=0.6$ and top-$p=0.95$.
Our setup differs only in the maximum model sequence length; we use $65{,}536$ instead of $32{,}768$ because multilingual generations may require more tokens to express the same semantic content.
To reduce generation variation, models are evaluated using consistent inference settings across languages and difficulty levels with three seed values.

For each problem, we use \textbf{forced-target-language} prompting in which model is instructed to reason and answer in the target language. The prompting strategy is explained in Appendix~\ref{append:prompting_method}.

\paragraph{Evaluation Metrics.}
In addition to the standard accuracy reported on PolyMath, we evaluate thinking-language consistency (TC) and answer-language consistency (AC) scores.
Let $\hat{\ell}^{,r}$ denote the detected language of the reasoning trace and $\ell_i$ the requested language. We define TC as:
\begin{equation}
\text{TC} =
\frac{1}{N}\sum\mathbbm{1}[\hat{\ell}^{,r}_i = \ell_i].
\end{equation}

Similarly, let $\hat{\ell}^{,a}$ denote the detected language of the final answer. We define AC as:
\begin{equation}
\text{AC} =
\frac{1}{N}\sum\mathbbm{1}[\hat{\ell}^{,a}_i = \ell_i].
\end{equation}

We report overall and difficulty-wise accuracy, TC, and AC scores across languages and models to study whether increasing task difficulty affects final-answer correctness and language consistency differently. Further hardware and software details are reported in Appendix~\ref{append:hardware-software}.

\section{Results}
\label{sec:results}













\begin{figure*}[p]
\centering

\addtocounter{figure}{-1}

DeepSeek-R1-Distill-Qwen-7B


\vspace{0.25cm}
\begin{tikzpicture}
\begin{axis}[
    hide axis,
    xmin=0, xmax=1, ymin=0, ymax=1,
    width=15cm,
    height=6cm,
    legend columns=3,
    legend style={
        draw=black!30,
        fill=white,
        font=\small,
        at={(0.5,0.5)},
        anchor=center,
        column sep=0.2cm,
    },
]
\addlegendimage{
    color=green!60!black,
    dotted,
    mark=*,
    thick,
    mark size=2pt
}
\addlegendentry{Accuracy}

\addlegendimage{
    color=orange!90!black,
    solid,
    mark=*,
    thick,
    mark size=2pt
}
\addlegendentry{Answer Consistency (AC)}

\addlegendimage{
    color=red!85!black,
    solid,
    mark=*,
    thick,
    mark size=2pt
}
\addlegendentry{Thinking Consistency (TC)}

\end{axis}
\end{tikzpicture}

\vspace{0.4cm}

\begin{subfigure}[b]{0.24\textwidth}
\centering
\begin{tikzpicture}
\begin{axis}[
    width=\textwidth,
    height=0.9\textwidth,
    xlabel={Difficulty Level},
    ylabel={Score (\%)},
    xmin=0.5, xmax=4.5,
    ymin=-5, ymax=105,
    xtick={1,2,3,4},
    ytick={0,20,40,60,80,100},
    tick label style={font=\small},
    label style={font=\small},
    grid=major,
    grid style={line width=.1pt, draw=gray!30},
    major grid style={line width=.2pt, draw=gray!50},
]

\addplot[
    color=green!60!black,
    dotted,
    mark=*,
    thick,
    mark size=1.0pt
] coordinates {
    (1,80.0) (2,56.8) (3,39.5) (4,20.5)
};

\addplot[
    color=red!85!black,
    solid,
    mark=*,
    thick,
    mark size=1.0pt
] coordinates {
    (1,100.0) (2,100.0) (3,99.7) (4,99.5)
};

\addplot[
    color=orange!90!black,
    solid,
    mark=*,
    thick,
    mark size=1.0pt
] coordinates {
    (1,99.7) (2,100.0) (3,100.0) (4,100.0)
};

\end{axis}
\end{tikzpicture}
\caption{English}
\label{fig:en-ds7b-combined}
\end{subfigure}
\hfill
\begin{subfigure}[b]{0.24\textwidth}
\centering
\begin{tikzpicture}
\begin{axis}[
    width=\textwidth,
    height=0.9\textwidth,
    xlabel={Difficulty Level},
    xmin=0.5, xmax=4.5,
    ymin=-5, ymax=105,
    xtick={1,2,3,4},
    ytick={0,20,40,60,80,100},
    tick label style={font=\small},
    label style={font=\small},
    grid=major,
    grid style={line width=.1pt, draw=gray!30},
    major grid style={line width=.2pt, draw=gray!50},
]

\addplot[
    color=green!60!black,
    dotted,
    mark=*,
    thick,
    mark size=1.0pt
] coordinates {
    (1,80.0) (2,53.6) (3,32.3) (4,18.1)
};

\addplot[
    color=red!85!black,
    solid,
    mark=*,
    thick,
    mark size=1.0pt
] coordinates {
    (1,94.7) (2,93.5) (3,91.5) (4,92.7)
};

\addplot[
    color=orange!90!black,
    solid,
    mark=*,
    thick,
    mark size=1.0pt
] coordinates {
    (1,99.5) (2,95.5) (3,96.7) (4,94.9)
};

\end{axis}
\end{tikzpicture}
\caption{Chinese}
\label{fig:zh-ds7b-combined}
\end{subfigure}
\hfill
\begin{subfigure}[b]{0.24\textwidth}
\centering
\begin{tikzpicture}
\begin{axis}[
    width=\textwidth,
    height=0.9\textwidth,
    xlabel={Difficulty Level},
    xmin=0.5, xmax=4.5,
    ymin=-5, ymax=105,
    xtick={1,2,3,4},
    ytick={0,20,40,60,80,100},
    tick label style={font=\small},
    label style={font=\small},
    grid=major,
    grid style={line width=.1pt, draw=gray!30},
    major grid style={line width=.2pt, draw=gray!50},
]

\addplot[
    color=green!60!black,
    dotted,
    mark=*,
    thick,
    mark size=1.0pt
] coordinates {
    (1,53.9) (2,54.7) (3,31.7) (4,16.5)
};

\addplot[
    color=red!85!black,
    solid,
    mark=*,
    thick,
    mark size=1.0pt
] coordinates {
    (1,70.1) (2,3.7) (3,4.3) (4,2.9)
};

\addplot[
    color=orange!90!black,
    solid,
    mark=*,
    thick,
    mark size=1.0pt
] coordinates {
    (1,75.7) (2,4.0) (3,2.1) (4,1.1)
};

\end{axis}
\end{tikzpicture}
\caption{Arabic}
\label{fig:ar-ds7b-combined}
\end{subfigure}
\hfill
\begin{subfigure}[b]{0.24\textwidth}
\centering
\begin{tikzpicture}
\begin{axis}[
    width=\textwidth,
    height=0.9\textwidth,
    xlabel={Difficulty Level},
    xmin=0.5, xmax=4.5,
    ymin=-5, ymax=105,
    xtick={1,2,3,4},
    ytick={0,20,40,60,80,100},
    tick label style={font=\small},
    label style={font=\small},
    grid=major,
    grid style={line width=.1pt, draw=gray!30},
    major grid style={line width=.2pt, draw=gray!50},
]

\addplot[
    color=green!60!black,
    dotted,
    mark=*,
    thick,
    mark size=1.0pt
] coordinates {
    (1,47.5) (2,50.9) (3,30.4) (4,16.5)
};

\addplot[
    color=red!85!black,
    solid,
    mark=*,
    thick,
    mark size=1.0pt
] coordinates {
    (1,97.9) (2,1.6) (3,2.1) (4,3.2)
};

\addplot[
    color=orange!90!black,
    solid,
    mark=*,
    thick,
    mark size=1.0pt
] coordinates {
    (1,98.1) (2,0.5) (3,1.3) (4,2.9)
};

\end{axis}
\end{tikzpicture}
\caption{Bengali}
\label{fig:bn-ds7b-combined}
\end{subfigure}

\vspace{0.5cm}

\begin{subfigure}[b]{0.24\textwidth}
\centering
\begin{tikzpicture}
\begin{axis}[
    width=\textwidth,
    height=0.9\textwidth,
    xlabel={Difficulty Level},
    ylabel={Score (\%)},
    xmin=0.5, xmax=4.5,
    ymin=-5, ymax=105,
    xtick={1,2,3,4},
    ytick={0,20,40,60,80,100},
    tick label style={font=\small},
    label style={font=\small},
    grid=major,
    grid style={line width=.1pt, draw=gray!30},
    major grid style={line width=.2pt, draw=gray!50},
]

\addplot[
    color=green!60!black,
    dotted,
    mark=*,
    thick,
    mark size=1.0pt
] coordinates {
    (1,60.5) (2,58.7) (3,32.5) (4,16.5)
};

\addplot[
    color=red!85!black,
    solid,
    mark=*,
    thick,
    mark size=1.0pt
] coordinates {
    (1,100.0) (2,19.5) (3,22.4) (4,9.3)
};

\addplot[
    color=orange!90!black,
    solid,
    mark=*,
    thick,
    mark size=1.0pt
] coordinates {
    (1,100.0) (2,18.7) (3,17.9) (4,6.9)
};

\end{axis}
\end{tikzpicture}
\caption{French}
\label{fig:fr-ds7b-combined}
\end{subfigure}
\hfill
\begin{subfigure}[b]{0.24\textwidth}
\centering
\begin{tikzpicture}
\begin{axis}[
    width=\textwidth,
    height=0.9\textwidth,
    xlabel={Difficulty Level},
    xmin=0.5, xmax=4.5,
    ymin=-5, ymax=105,
    xtick={1,2,3,4},
    ytick={0,20,40,60,80,100},
    tick label style={font=\small},
    label style={font=\small},
    grid=major,
    grid style={line width=.1pt, draw=gray!30},
    major grid style={line width=.2pt, draw=gray!50},
]

\addplot[
    color=green!60!black,
    dotted,
    mark=*,
    thick,
    mark size=1.0pt
] coordinates {
    (1,61.3) (2,49.6) (3,26.4) (4,13.1)
};

\addplot[
    color=red!85!black,
    solid,
    mark=*,
    thick,
    mark size=1.0pt
] coordinates {
    (1,97.1) (2,85.3) (3,86.7) (4,86.7)
};

\addplot[
    color=orange!90!black,
    solid,
    mark=*,
    thick,
    mark size=1.0pt
] coordinates {
    (1,95.7) (2,67.5) (3,53.6) (4,53.3)
};

\end{axis}
\end{tikzpicture}
\caption{Russian}
\label{fig:ru-ds7b-combined}
\end{subfigure}
\hfill
\begin{subfigure}[b]{0.24\textwidth}
\centering
\begin{tikzpicture}
\begin{axis}[
    width=\textwidth,
    height=0.9\textwidth,
    xlabel={Difficulty Level},
    xmin=0.5, xmax=4.5,
    ymin=-5, ymax=105,
    xtick={1,2,3,4},
    ytick={0,20,40,60,80,100},
    tick label style={font=\small},
    label style={font=\small},
    grid=major,
    grid style={line width=.1pt, draw=gray!30},
    major grid style={line width=.2pt, draw=gray!50},
]

\addplot[
    color=green!60!black,
    dotted,
    mark=*,
    thick,
    mark size=1.0pt
] coordinates {
    (1,4.5) (2,35.7) (3,17.1) (4,6.9)
};

\addplot[
    color=red!85!black,
    solid,
    mark=*,
    thick,
    mark size=1.0pt
] coordinates {
    (1,89.9) (2,1.9) (3,7.2) (4,4.8)
};

\addplot[
    color=orange!90!black,
    solid,
    mark=*,
    thick,
    mark size=1.0pt
] coordinates {
    (1,85.3) (2,1.6) (3,5.1) (4,3.7)
};

\end{axis}
\end{tikzpicture}
\caption{Swahili}
\label{fig:sw-ds7b-combined}
\end{subfigure}
\hfill
\begin{subfigure}[b]{0.24\textwidth}
\centering
\begin{tikzpicture}
\begin{axis}[
    width=\textwidth,
    height=0.9\textwidth,
    xlabel={Difficulty Level},
    xmin=0.5, xmax=4.5,
    ymin=-5, ymax=105,
    xtick={1,2,3,4},
    ytick={0,20,40,60,80,100},
    tick label style={font=\small},
    label style={font=\small},
    grid=major,
    grid style={line width=.1pt, draw=gray!30},
    major grid style={line width=.2pt, draw=gray!50},
]

\addplot[
    color=green!60!black,
    dotted,
    mark=*,
    thick,
    mark size=1.0pt
] coordinates {
    (1,22.9) (2,41.9) (3,24.0) (4,12.5)
};

\addplot[
    color=red!85!black,
    solid,
    mark=*,
    thick,
    mark size=1.0pt
] coordinates {
    (1,99.2) (2,11.7) (3,9.3) (4,4.3)
};

\addplot[
    color=orange!90!black,
    solid,
    mark=*,
    thick,
    mark size=1.0pt
] coordinates {
    (1,96.8) (2,8.0) (3,7.5) (4,4.3)
};

\end{axis}
\end{tikzpicture}
\caption{Telugu}
\label{fig:te-ds7b-combined}
\end{subfigure}

\captionof{figure}{
Difficulty-wise final-answer accuracy, thinking-language consistency (TC),
and answer-language consistency (AC) of DeepSeek-R1-Distill-Qwen-7B across eight languages and four PolyMath difficulty levels: 1 = low, 2 = medium, 3 = high, and 4 = top.
Results: accuracy generally decreases with task-difficulty, but thinking- and answer-language consistency are language-dependent: accurate in English and Chinese, but breaking down for languages like Arabic, Bengali, French, Swahili, and Telugu.
}
\label{fig:bf16_accuracy_tc_ac_difficulty_ds7b}


\vspace{0.4cm}

OLMo-3-7B-Think


\vspace{0.25cm}

\begin{tikzpicture}
\begin{axis}[
    hide axis,
    xmin=0, xmax=1, ymin=0, ymax=1,
    width=15cm,
    height=6cm,
    legend columns=3,
    legend style={
        draw=black!30,
        fill=white,
        font=\small,
        at={(0.5,0.5)},
        anchor=center,
        column sep=0.2cm,
    },
]
\addlegendimage{
    color=green!60!black,
    dotted,
    mark=*,
    thick,
    mark size=2pt
}
\addlegendentry{Accuracy}

\addlegendimage{
    color=orange!90!black,
    solid,
    mark=*,
    thick,
    mark size=2pt
}
\addlegendentry{Answer Consistency (AC)}

\addlegendimage{
    color=red!85!black,
    solid,
    mark=*,
    thick,
    mark size=2pt
}
\addlegendentry{Thinking Consistency (TC)}

\end{axis}
\end{tikzpicture}

\vspace{0.4cm}

\begin{subfigure}[b]{0.24\textwidth}
\centering
\begin{tikzpicture}
\begin{axis}[
    width=\textwidth,
    height=0.9\textwidth,
    xlabel={Difficulty Level},
    ylabel={Score (\%)},
    xmin=0.5, xmax=4.5,
    ymin=-5, ymax=105,
    xtick={1,2,3,4},
    ytick={0,20,40,60,80,100},
    tick label style={font=\small},
    label style={font=\small},
    grid=major,
    grid style={line width=.1pt, draw=gray!30},
    major grid style={line width=.2pt, draw=gray!50},
]

\addplot[
    color=green!60!black,
    dotted,
    mark=*,
    thick,
    mark size=1.0pt
] coordinates {
    (1,92.5) (2,63.2) (3,51.5) (4,23.2)
};

\addplot[
    color=orange!85!black,
    solid,
    mark=*,
    thick,
    mark size=1.0pt
] coordinates {
    (1,100.0) (2,100.0) (3,100.0) (4,98.9)
};

\addplot[
    color=red!90!black,
    solid,
    mark=*,
    thick,
    mark size=1.0pt
] coordinates {
    (1,99.7) (2,100.0) (3,100.0) (4,100.0)
};

\end{axis}
\end{tikzpicture}
\caption{English}
\label{fig:en-olmo7b-combined}
\end{subfigure}
\hfill
\begin{subfigure}[b]{0.24\textwidth}
\centering
\begin{tikzpicture}
\begin{axis}[
    width=\textwidth,
    height=0.9\textwidth,
    xlabel={Difficulty Level},
    xmin=0.5, xmax=4.5,
    ymin=-5, ymax=105,
    xtick={1,2,3,4},
    ytick={0,20,40,60,80,100},
    tick label style={font=\small},
    label style={font=\small},
    grid=major,
    grid style={line width=.1pt, draw=gray!30},
    major grid style={line width=.2pt, draw=gray!50},
]

\addplot[
    color=green!60!black,
    dotted,
    mark=*,
    thick,
    mark size=1.0pt
] coordinates {
    (1,79.2) (2,61.9) (3,51.2) (4,23.7)
};

\addplot[
    color=red!85!black,
    solid,
    mark=*,
    thick,
    mark size=1.0pt
] coordinates {
    (1,18.9) (2,11.2) (3,7.5) (4,7.7)
};

\addplot[
    color=orange!90!black,
    solid,
    mark=*,
    thick,
    mark size=1.0pt
] coordinates {
    (1,77.6) (2,13.9) (3,6.7) (4,7.2)
};

\end{axis}
\end{tikzpicture}
\caption{Chinese}
\label{fig:zh-olmo7b-combined}
\end{subfigure}
\hfill
\begin{subfigure}[b]{0.24\textwidth}
\centering
\begin{tikzpicture}
\begin{axis}[
    width=\textwidth,
    height=0.9\textwidth,
    xlabel={Difficulty Level},
    xmin=0.5, xmax=4.5,
    ymin=-5, ymax=105,
    xtick={1,2,3,4},
    ytick={0,20,40,60,80,100},
    tick label style={font=\small},
    label style={font=\small},
    grid=major,
    grid style={line width=.1pt, draw=gray!30},
    major grid style={line width=.2pt, draw=gray!50},
]

\addplot[
    color=green!60!black,
    dotted,
    mark=*,
    thick,
    mark size=1.0pt
] coordinates {
    (1,67.7) (2,55.2) (3,34.7) (4,21.1)
};

\addplot[
    color=red!85!black,
    solid,
    mark=*,
    thick,
    mark size=1.0pt
] coordinates {
    (1,0.0) (2,0.5) (3,0.5) (4,1.1)
};

\addplot[
    color=orange!90!black,
    solid,
    mark=*,
    thick,
    mark size=1.0pt
] coordinates {
    (1,74.4) (2,1.3) (3,0.5) (4,1.6)
};

\end{axis}
\end{tikzpicture}
\caption{Arabic}
\label{fig:ar-olmo7b-combined}
\end{subfigure}
\hfill
\begin{subfigure}[b]{0.24\textwidth}
\centering
\begin{tikzpicture}
\begin{axis}[
    width=\textwidth,
    height=0.9\textwidth,
    xlabel={Difficulty Level},
    xmin=0.5, xmax=4.5,
    ymin=-5, ymax=105,
    xtick={1,2,3,4},
    ytick={0,20,40,60,80,100},
    tick label style={font=\small},
    label style={font=\small},
    grid=major,
    grid style={line width=.1pt, draw=gray!30},
    major grid style={line width=.2pt, draw=gray!50},
]

\addplot[
    color=green!60!black,
    dotted,
    mark=*,
    thick,
    mark size=1.0pt
] coordinates {
    (1,44.8) (2,45.9) (3,28.5) (4,12.3)
};

\addplot[
    color=red!85!black,
    solid,
    mark=*,
    thick,
    mark size=1.0pt
] coordinates {
    (1,1.9) (2,2.1) (3,3.5) (4,5.1)
};

\addplot[
    color=orange!90!black,
    solid,
    mark=*,
    thick,
    mark size=1.0pt
] coordinates {
    (1,51.5) (2,0.8) (3,0.8) (4,1.9)
};

\end{axis}
\end{tikzpicture}
\caption{Bengali}
\label{fig:bn-olmo7b-combined}
\end{subfigure}

\vspace{0.5cm}

\begin{subfigure}[b]{0.24\textwidth}
\centering
\begin{tikzpicture}
\begin{axis}[
    width=\textwidth,
    height=0.9\textwidth,
    xlabel={Difficulty Level},
    ylabel={Score (\%)},
    xmin=0.5, xmax=4.5,
    ymin=-5, ymax=105,
    xtick={1,2,3,4},
    ytick={0,20,40,60,80,100},
    tick label style={font=\small},
    label style={font=\small},
    grid=major,
    grid style={line width=.1pt, draw=gray!30},
    major grid style={line width=.2pt, draw=gray!50},
]

\addplot[
    color=green!60!black,
    dotted,
    mark=*,
    thick,
    mark size=1.0pt
] coordinates {
    (1,74.9) (2,62.7) (3,52.5) (4,25.3)
};

\addplot[
    color=red!85!black,
    solid,
    mark=*,
    thick,
    mark size=1.0pt
] coordinates {
    (1,0.3) (2,0.5) (3,0.3) (4,0.3)
};

\addplot[
    color=orange!90!black,
    solid,
    mark=*,
    thick,
    mark size=1.0pt
] coordinates {
    (1,95.2) (2,13.3) (3,5.9) (4,8.3)
};

\end{axis}
\end{tikzpicture}
\caption{French}
\label{fig:fr-olmo7b-combined}
\end{subfigure}
\hfill
\begin{subfigure}[b]{0.24\textwidth}
\centering
\begin{tikzpicture}
\begin{axis}[
    width=\textwidth,
    height=0.9\textwidth,
    xlabel={Difficulty Level},
    xmin=0.5, xmax=4.5,
    ymin=-5, ymax=105,
    xtick={1,2,3,4},
    ytick={0,20,40,60,80,100},
    tick label style={font=\small},
    label style={font=\small},
    grid=major,
    grid style={line width=.1pt, draw=gray!30},
    major grid style={line width=.2pt, draw=gray!50},
]

\addplot[
    color=green!60!black,
    dotted,
    mark=*,
    thick,
    mark size=1.0pt
] coordinates {
    (1,66.1) (2,46.4) (3,28.3) (4,11.7)
};

\addplot[
    color=red!85!black,
    solid,
    mark=*,
    thick,
    mark size=1.0pt
] coordinates {
    (1,100.0) (2,98.7) (3,99.7) (4,99.2)
};

\addplot[
    color=orange!90!black,
    solid,
    mark=*,
    thick,
    mark size=1.0pt
] coordinates {
    (1,98.9) (2,80.0) (3,86.1) (4,88.3)
};

\end{axis}
\end{tikzpicture}
\caption{Russian}
\label{fig:ru-olmo7b-combined}
\end{subfigure}
\hfill
\begin{subfigure}[b]{0.24\textwidth}
\centering
\begin{tikzpicture}
\begin{axis}[
    width=\textwidth,
    height=0.9\textwidth,
    xlabel={Difficulty Level},
    xmin=0.5, xmax=4.5,
    ymin=-5, ymax=105,
    xtick={1,2,3,4},
    ytick={0,20,40,60,80,100},
    tick label style={font=\small},
    label style={font=\small},
    grid=major,
    grid style={line width=.1pt, draw=gray!30},
    major grid style={line width=.2pt, draw=gray!50},
]

\addplot[
    color=green!60!black,
    dotted,
    mark=*,
    thick,
    mark size=1.0pt
] coordinates {
    (1,21.1) (2,35.5) (3,22.9) (4,12.3)
};

\addplot[
    color=red!85!black,
    solid,
    mark=*,
    thick,
    mark size=1.0pt
] coordinates {
    (1,0.5) (2,0.5) (3,0.0) (4,1.6)
};

\addplot[
    color=orange!90!black,
    solid,
    mark=*,
    thick,
    mark size=1.0pt
] coordinates {
    (1,43.2) (2,1.6) (3,2.1) (4,3.5)
};

\end{axis}
\end{tikzpicture}
\caption{Swahili}
\label{fig:sw-olmo7b-combined}
\end{subfigure}
\hfill
\begin{subfigure}[b]{0.24\textwidth}
\centering
\begin{tikzpicture}
\begin{axis}[
    width=\textwidth,
    height=0.9\textwidth,
    xlabel={Difficulty Level},
    xmin=0.5, xmax=4.5,
    ymin=-5, ymax=105,
    xtick={1,2,3,4},
    ytick={0,20,40,60,80,100},
    tick label style={font=\small},
    label style={font=\small},
    grid=major,
    grid style={line width=.1pt, draw=gray!30},
    major grid style={line width=.2pt, draw=gray!50},
]

\addplot[
    color=green!60!black,
    dotted,
    mark=*,
    thick,
    mark size=1.0pt
] coordinates {
    (1,50.7) (2,48.8) (3,34.1) (4,16.8)
};

\addplot[
    color=red!85!black,
    solid,
    mark=*,
    thick,
    mark size=1.0pt
] coordinates {
    (1,0.0) (2,1.1) (3,1.9) (4,1.6)
};

\addplot[
    color=orange!90!black,
    solid,
    mark=*,
    thick,
    mark size=1.0pt
] coordinates {
    (1,36.3) (2,0.0) (3,0.3) (4,0.3)
};

\end{axis}
\end{tikzpicture}
\caption{Telugu}
\label{fig:te-olmo7b-combined}
\end{subfigure}

\captionof{figure}{
Analogous plot to Figure \ref{fig:bf16_accuracy_tc_ac_difficulty_ds7b}, for OLMo-3-7B-Think. English and Russian maintain language consistency, while other languages consistency degrades rapidly with task difficulty.
}
\label{fig:bf16_accuracy_tc_ac_difficulty_olmo7b}

\end{figure*}

\begin{table*}[!t]
\centering
\renewcommand{\arraystretch}{1.20}
\resizebox{0.92\linewidth}{!}{
\begin{tabular}{llcccll}
\hline\hline
\textbf{Model} &
\textbf{Method} &
\textbf{W-A-KV Bits} &
\textbf{Avg. $\Delta$Acc.} &
\textbf{Avg. $\Delta$TC} &
\textbf{TC Co-wins} &
\textbf{TC Co-winner Lang.} \\
\hline

\multirow{3}{*}{\textbf{\shortstack[l]{DeepSeek-R1\\Distill-Qwen-7B}}}
& GPTQ & 4-16-16 &
\cellcolor{red!15}-3.39 &
-5.40 &
\cellcolor{yellow!15}{6/8} &
\cellcolor{yellow!15}{en, zh, ar, fr, sw, te} \\

& AWQ & 4-16-16 &
-2.53 &
\cellcolor{green!15}-3.83 &
5/8 &
en, ar, bn, fr, ru \\

& AutoRound & 4-16-16 &
\cellcolor{green!15}-0.80 &
\cellcolor{red!15}-6.75 &
3/8 &
en, ar, bn \\
\hline

\multirow{3}{*}{\textbf{OLMo-3-7B-Think}}
& GPTQ & 4-16-16 &
-2.33 &
\cellcolor{green!15}+4.28 &
\cellcolor{yellow!15}{7/8} &
\cellcolor{yellow!15}{en, zh, ar, bn, ru, sw, te} \\

& AWQ & 4-16-16 &
\cellcolor{red!15}-2.39 &
+2.04 &
6/8 &
en, ar, bn, fr, ru, sw \\

& AutoRound & 4-16-16 &
\cellcolor{green!15}-1.14 &
\cellcolor{red!15}-0.19 &
5/8 &
en, ar, fr, sw, te \\
\hline\hline
\end{tabular}
}
\caption{
Summary of accuracy and thinking-language consistency (TC) $\Delta$ scores
relative to Baseline across eight languages for DeepSeek-R1-Distill-Qwen-7B and
OLMo-3-7B-Think.
Avg. $\Delta$Acc. and Avg. $\Delta$TC report average changes over the eight
shown languages.
Green/red cells mark the best/worst average values among GPTQ, AWQ, and
AutoRound within each model.
Yellow cells summarize tolerance-based TC voting co-winners with
$\epsilon=1.0$, i.e., methods within one TC point of the best $\Delta$TC for
each model-language group.
Full per-language Acc./TC and $\Delta$Acc./$\Delta$TC results are reported in
Table~\ref{tab:quant-acc-tc-scores-compact},~\ref{tab:quant-acc-tc-scores}, and ~\ref{tab:quant-acc-tc-deltas}.
}
\vspace{-2ex}
\label{tab:acc-tc-deltas-compact}
\end{table*}

 Table~\ref{tab:main-table-difficulty_breakdown} summarizes these difficulty-dependent behaviors across five reasoning and three non-reasoning models. Figures~\ref{fig:bf16_accuracy_tc_ac_difficulty_ds7b} and~\ref{fig:bf16_accuracy_tc_ac_difficulty_olmo7b} report final-answer accuracy, thinking-language consistency (TC), and answer-language consistency (AC) across four difficulty levels for DeepSeek-R1-Distill-Qwen-7B and OLMo-3-7B-Think, respectively. Results for Qwen3-30B-A3B-Instruct, Qwen3-30B-A3B-Thinking, DeepSeek-R1-Distill-Qwen-1.5B, Gemma-3-12B-IT, Phi-4-mini-instruct, and Phi-4-mini-reasoning are provided in Appendix Figures~\ref{fig:bf16_accuracy_tc_ac_difficulty_qwen3_30b_a3b_instruct},
\ref{fig:bf16_accuracy_tc_ac_difficulty_qwen3_30b_a3b_thinking},
\ref{fig:bf16_accuracy_tc_ac_difficulty_deepseek_r1_distill_qwen_1p5b},
\ref{fig:bf16_accuracy_tc_ac_difficulty_gemma3_12b_it},
\ref{fig:bf16_accuracy_tc_ac_difficulty_phi4mini_instruct}, and
\ref{fig:bf16_accuracy_tc_ac_difficulty_phi4mini_reasoning}.

\subsection{Four Difficulty-Dependent Language-Consistency Cases}
\label{sec:results-four-behaviors}

Output-language consistency does not follow a single pattern as task difficulty increases. Across models and languages, we identify four representative cases based on how the output language aligns with the input language from low to top difficulty, as illustrated in Figure~\ref{fig:cases-language-consistency-phenomena-cases}.

\textbf{Case 1: Output language remains consistent with the input language.}
Some models preserve the requested language across all difficulty levels, even as task accuracy decreases. For example, Phi-4-mini-reasoning maintains approximately 100\% answer- and thinking-language consistency for English from low through top difficulty, while accuracy decreases from 94.0\% to 14.0\% (Figure~\ref{fig:bf16_accuracy_tc_ac_difficulty_phi4mini_reasoning}).

\textbf{Case 2: Output language remains inconsistent with the input language.}
Some models consistently reason in a language different from the requested input language. For Arabic, Qwen3-30B-A3B-Thinking-2507 maintains 0\% thinking-language consistency across all difficulty levels, indicating that model prefers to use dominant internal reasoning language (English) despite the Arabic input. Similar behavior appears for Bengali, French, Swahili, and Telugu (Figure~\ref{fig:bf16_accuracy_tc_ac_difficulty_qwen3_30b_a3b_thinking}).

\textbf{Case 3: Output-language consistency declines gradually.}
Language consistency may progressively decrease as task difficulty increases. For Russian in DeepSeek-R1-Distill-Qwen-7B, thinking-language consistency decreases from 97.1\% at low difficulty to 86.7\% at top difficulty (Figure~\ref{fig:bf16_accuracy_tc_ac_difficulty_ds7b}).

\textbf{Case 4: Output-language consistency declines abruptly.}
In some cases, language consistency exhibits a sudden and substantial decrease between consecutive difficulty levels. For Telugu, DeepSeek-R1-Distill-Qwen-7B decreases from 99.2\% at low difficulty to 11.7\% at medium difficulty (Figure~\ref{fig:bf16_accuracy_tc_ac_difficulty_ds7b}). We formally define and analyze this behavior as the \textit{language consistency breakdown effect} in Section~\ref{sec:results-language-consistency-breakdown}.

\subsection{Language Consistency Breakdown Effect}
\label{sec:results-language-consistency-breakdown}

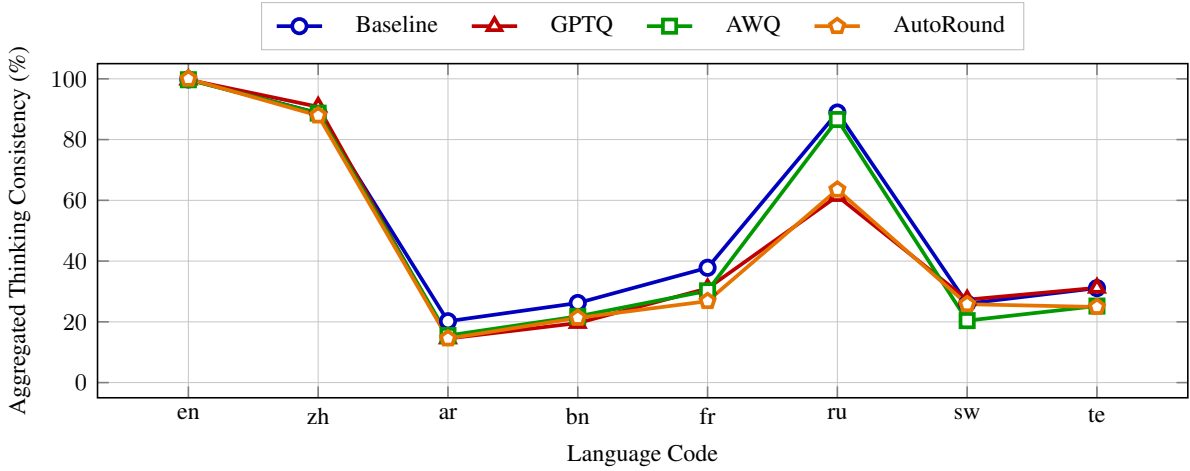
\begin{figure*}[t]
\centering
\begin{tikzpicture}
\begin{axis}[
    width=\textwidth,
    height=6.0cm,
    xlabel={Language Code},
    ylabel={Aggregated Thinking Consistency (\%)},
    ymin=-5, ymax=105,
    xtick=data,
    symbolic x coords={en,zh,ar,bn,fr,ru,sw,te},
    ytick={0,20,40,60,80,100},
    tick label style={font=\small},
    label style={font=\small},
    axis line style={line width=0.6pt},
    tick style={line width=0.6pt},
    grid=major,
    major grid style={
        line width=0.3pt,
        draw=gray!45
    },
    legend style={
        at={(0.5,1.04)},
        anchor=south,
        legend columns=4,
        draw=black!25,
        fill=white,
        font=\small,
        column sep=0.35cm,
        inner xsep=5pt,
        inner ysep=3pt
    },
]

\addplot[
    color=blue!75!black,
    solid,
    mark=*,
    mark options={fill=white, solid},
    line width=1.4pt,
    mark size=2.7pt
] coordinates {
    (en,99.8)
    (zh,88.7)
    (ar,20.2)
    (bn,26.2)
    (fr,37.8)
    (ru,88.9)
    (sw,26.0)
    (te,31.1)
};
\addlegendentry{Baseline}

\addplot[
    color=red!75!black,
    solid,
    mark=triangle*,
    mark options={fill=white, solid},
    line width=1.4pt,
    mark size=3.0pt
] coordinates {
    (en,99.7)
    (zh,90.8)
    (ar,14.5)
    (bn,19.6)
    (fr,31.0)
    (ru,61.4)
    (sw,27.3)
    (te,31.2)
};
\addlegendentry{GPTQ}

\addplot[
    color=green!60!black,
    solid,
    mark=square*,
    mark options={fill=white, solid},
    line width=1.4pt,
    mark size=2.5pt
] coordinates {
    (en,99.7)
    (zh,88.7)
    (ar,15.5)
    (bn,21.8)
    (fr,30.2)
    (ru,86.6)
    (sw,20.4)
    (te,25.2)
};
\addlegendentry{AWQ}

\addplot[
    color=orange!90!black,
    solid,
    mark=pentagon*,
    mark options={fill=white, solid},
    line width=1.4pt,
    mark size=2.8pt
] coordinates {
    (en,100.0)
    (zh,87.9)
    (ar,14.5)
    (bn,21.3)
    (fr,26.8)
    (ru,63.5)
    (sw,25.8)
    (te,24.9)
};
\addlegendentry{AutoRound}

\end{axis}
\end{tikzpicture}

\caption{
Aggregated thinking-language consistency of Baseline and W4A16-quantized
DeepSeek-R1-Distill-Qwen-7B across eight languages.
}
\label{fig:flow-ds7b_quantized_tc_eight_languages}
\end{figure*}

As task difficulty increases, reasoning models often become less consistent in the language used for their thinking traces and final answers. We define the \textit{language consistency breakdown effect} using a user-defined drop threshold $\Delta_{\text{break}}$. Between consecutive difficulty levels $d$ and $d+1$, a breakdown occurs when
\vspace{-0.1cm}
\begin{equation}
\Delta TC_d = TC_d - TC_{d+1} \geq \Delta_{\text{break}},
\end{equation}
\begin{equation}
\Delta AC_d = AC_d - AC_{d+1} \geq \Delta_{\text{break}}.
\label{eq:language-consistency-breakdown}
\end{equation}

We set $\Delta_{\text{break}}=30$ percentage points and report sensitivity results for $\Delta_{break} \in \{30, 50, 70\}$ in Appendix~\ref{sec:breakdown-threshold-sensitivity}. For DeepSeek-R1-Distill-Qwen-7B, Bengali TC drops from 97.9\% to 1.6\%, while AC drops from 98.1\% to 0.5\% between low and medium difficulty. Similar breakdowns occur for Arabic, French, Swahili, and Telugu, whereas Chinese and Russian remain comparatively stable.

The effect is strongly model-dependent. DeepSeek-R1-Distill-Qwen-1.5B and 7B exhibit AC and TC breakdowns in 5/8 languages, respectively.  Qwen3-30B-A3B-Thinking-2507 exhibits no breakdown in either metric.

\textbf{Increasing task difficulty generally reduces TC and AC, with some languages exhibiting abrupt breakdowns in one or both.}






\subsection{Language consistency breakdown breaks monotonic accuracy--difficulty trend}
\label{sec:results-nonlinear-breakdown}

Model accuracy is generally expected to decline monotonically as task difficulty increases. However, language consistency breakdown effect introduces non-linear behavior that breaks this expected relationship: accuracy can preserve or improve at a higher difficulty level when TC and AC collapse.

For DeepSeek-R1-Distill-Qwen-7B, Bengali accuracy increases (Figure~\ref{fig:bf16_accuracy_tc_ac_difficulty_ds7b}) from 47.5\% at low difficulty to 50.9\% at medium difficulty, while TC decreases from 97.9\% to 1.6\% and AC decreases from 98.1\% to 0.5\%. Swahili accuracy similarly increases from 4.5\% to 35.7\%, despite TC and AC collapsing from 89.9\% and 85.3\% to 1.9\% and 1.6\%, respectively. Telugu exhibits the same pattern, with accuracy increasing from 22.9\% to 41.9\% while TC and AC decrease sharply.

These results show that improved accuracy at a harder difficulty level does not necessarily indicate improved reasoning or answer consistency. Instead, the model may solve more problems by shifting toward a more dominant internal language during thinking or answer generation.

\textbf{The language consistency breakdown breaks the expected monotonic accuracy--difficulty relationship, revealing that accuracy can preserve or increase at a higher difficulty level when language consistency collapses.}

\section{Analyses and Discussion}
\label{sec:analysis_discussion}


In this section, we analyze output-language consistency from perspectives of quantization, reasoning versus non-reasoning, and prompt control.

\subsection{Impact of Quantization}
\label{sec:analysis_quantization}

\textit{How does quantization affect multilingual reasoning accuracy and thinking-language consistency?}

Quantization affects multilingual reasoning unevenly: final-answer accuracy is generally preserved or reduced, whereas output-language consistency can either improve or degrade depending on the model, language, and quantization method. To the best of our knowledge, this is the first systematic study of how quantization affects thinking-language consistency in multilingual reasoning models.

Among the evaluated W4A16 integer methods, AutoRound provides the strongest accuracy preservation. For DeepSeek-R1-Distill-Qwen-7B, AutoRound reduces average accuracy by only $0.80$ points, compared with $2.53$ for AWQ and $3.39$ for GPTQ. Similarly, for OLMo-3-7B-Think, AutoRound yields the smallest degradation at $1.14$ points, compared with $2.33$ for GPTQ and $2.39$ for AWQ as showed in Table~\ref{tab:acc-tc-deltas-compact}. This gives the overall accuracy-preservation order:
\[
\text{AutoRound} \;>\; \text{AWQ} \;\geq\; \text{GPTQ}.
\]

This ordering does not extend to output-language consistency, since quantization affects language consistency differently. It can degrade or improve and can be skewed by large gains or drops in a small number of languages. 
Therefore, we use tolerance-based voting as the main criteria to rank quantization methods for TC and report avg.~$\Delta$TC as a complementary magnitude measure. The tolerance parameter $\epsilon$ defines the minimum TC difference considered meaningful, preventing very small gaps from being over-interpreted. A method is counted as a co-winner if
\[
\Delta TC_m \geq \max_{m' \in \mathcal{M}} \Delta TC_{m'} - \epsilon,
\]
where $\mathcal{M}=\{\text{GPTQ},\text{AWQ},\text{AutoRound}\}$ and $\epsilon=1.0$ percentage point. Under this criterion, GPTQ receives the better support for preserving or improving TC, while AWQ achieves larger-magnitude improvements; AutoRound performs comparatively worse as illustrated in Figure~\ref{fig:flow-ds7b_quantized_tc_eight_languages} and \ref{fig:flow-olmo_quantized_tc_eight_languages}. Appendix~\ref{append:tolerance-based-voting} reports additional TC voting results for $\epsilon \in \{0.5,1.0,2.0\}$.

\textbf{Quantization affects accuracy and language consistency differently: AutoRound best preserves final-answer accuracy, while GPTQ and AWQ more often preserve or improve thinking-language consistency.
}

\subsection{Reasoning versus Non-Reasoning}

\textit{How does task difficulty affect language consistency in reasoning and non-reasoning models?}

Reasoning and non-reasoning models exhibit distinct language-consistency behavior as task difficulty increases. The evaluated non-reasoning models remain comparatively stable, with no answer-language consistency (AC) breakdown across the eight languages. In contrast, DeepSeek-R1-Distill-Qwen-1.5B and 7B, OLMo-3-7B-Think, and Phi-4-mini-reasoning exhibit AC breakdowns in multiple languages as illustrated in Figure~\ref{fig:bf16_accuracy_tc_ac_difficulty_deepseek_r1_distill_qwen_1p5b}, \ref{fig:bf16_accuracy_tc_ac_difficulty_ds7b}, \ref{fig:bf16_accuracy_tc_ac_difficulty_olmo7b}, and \ref{fig:bf16_accuracy_tc_ac_difficulty_phi4mini_reasoning}.

\textbf{Language-consistency degradation becomes more pronounced in reasoning models as task difficulty increases.}

\subsection{Prompt Control}

\textit{Can allowing models to reason in their preferred language improve answer-language consistency?}

Our main experiments explicitly instruct each model to reason and answer in the target language. As a prompt control, we relax the reasoning-language constraint and allow the model to select its preferred reasoning language while still requiring the final answer in the target language. 

We evaluate this control on DeepSeek-R1-Distill-Qwen-7B and Phi-4-mini-reasoning. Allowing unrestricted reasoning does not ensure answer-language consistency. DeepSeek-R1-Distill-Qwen-7B achieves answer-language consistency below $\tau_{\text{cons}}=90\%$ at Top difficulty for 6/8 languages. Phi-4-mini-reasoning achieves answer-language consistency below the threshold for 7/8 languages. English remains highly consistent, whereas Arabic, Bengali, French, Swahili, and Telugu exhibit substantially lower AC (Table~\ref{tab:preferred-language-prompt-control}).

\textbf{Allowing models to reason in their preferred language does not reliably preserve the requested answer language, particularly at higher difficulty levels.}

\section{Conclusion}
\label{sec:conclusion}

In this paper, we uncover hidden language-consistency phenomena in reasoning and non-reasoning LLMs, revealing how task difficulty  affects final-answer accuracy, thinking-language consistency (TC), and answer-language consistency (AC). Using PolyMath across eight languages and four difficulty levels, we show that accuracy and language consistency capture distinct aspects of multilingual capability.

\looseness=-1
Our results reveal four difficulty-dependent language-consistency behaviors, ranging from consistent TC and AC to gradual degradation and abrupt collapse. Increasing task difficulty generally reduces thinking- and answer-language consistency, with some languages exhibiting the \textit{language consistency breakdown effect}, where TC or AC drops abruptly. This effect is more prevalent in lower-resource and non-Latin-script languages and is substantially more pronounced in several reasoning models, while the evaluated non-reasoning models maintain comparatively consistent answer-language consistency. We also uncover non-linear behavior that breaks the expected accuracy--difficulty relationship: accuracy can improve at a higher difficulty level as TC and AC collapse.

Quantization further affects these dimensions unevenly. It can improve or degrade language-consistency independently of task accuracy. Among the evaluated W4A16 integer methods, AutoRound best preserves final-answer accuracy, while AWQ and GPTQ more often preserve or improve thinking-language consistency under tolerance-based voting with $\epsilon=1.0$.

Overall, multilingual capability cannot be characterized by accuracy alone. Reliable evaluation should jointly consider task difficulty, accuracy, answer- and thinking-language consistency scores. 

\section{Limitations}
\label{sec:limitations}


\paragraph{Model coverage.}
{\sloppy \looseness=-1
We evaluate five open reasoning models and three non-reasoning models 
across several model families and scales. However, our findings may 
not generalize to substantially larger models, closed-source systems, 
or models trained with different reasoning and post-training strategies.\par}

\paragraph{Language coverage.}
We evaluate eight languages from PolyMath, selected to cover different scripts and levels of multilingual representation. The observed language-consistency patterns may differ for other languages, particularly those that are less represented in multilingual model training.

\paragraph{Benchmark coverage.}
Our experiments focus on PolyMath, a multilingual mathematical reasoning benchmark with four difficulty levels. The observed difficulty-related language-consistency breakdown may differ for other reasoning domains, such as scientific reasoning, coding, or open-ended tasks.

\paragraph{Prompt dependence.}
Our experiments use native-language reasoning and boxed-answer instructions. Different prompting strategies, such as English reasoning prompts across all languages, free-form responses, or few-shot prompting, may produce different language-consistency behavior.




\bibliography{custom}

\appendix

\begin{table*}[t]
\centering
\small
\begin{tabular}{l c l c c}
\toprule
\textbf{Model} & $\epsilon$ & \textbf{Method} & \textbf{Co-winner Ranking} & \textbf{Mean $\Delta$TC} \\
\midrule

\multirow{3}{*}{DeepSeek-R1-Distill-Qwen-1.5B} & \multirow{3}{*}{0.5}
& GPTQ      & 7 & 8.90 \\
& & AWQ       & 2 & 3.92 \\
& & AutoRound & 4 & 2.51 \\
\cline{1-5}

\multirow{3}{*}{DeepSeek-R1-Distill-Qwen-7B} & \multirow{3}{*}{0.5}
& GPTQ      & 5 & -5.40 \\
& & AWQ       & 4 & -3.83 \\
& & AutoRound & 2 & -6.75 \\
\cline{1-5}

\multirow{3}{*}{OLMo-3-7B-Think} & \multirow{3}{*}{0.5}
& GPTQ      & 5 & 4.28 \\
& & AWQ       & 7 & 2.04 \\
& & AutoRound & 1 & -0.19 \\
\hline \hline

\multirow{3}{*}{DeepSeek-R1-Distill-Qwen-1.5B} & \multirow{3}{*}{1.0}
& GPTQ      & 7 & 8.90 \\
& & AWQ       & 2 & 3.92 \\
& & AutoRound & 4 & 2.51 \\
\cline{1-5}

\multirow{3}{*}{DeepSeek-R1-Distill-Qwen-7B} & \multirow{3}{*}{1.0}
& GPTQ      & 6 & -5.40 \\
& & AWQ       & 5 & -3.83 \\
& & AutoRound & 3 & -6.75 \\
\cline{1-5}

\multirow{3}{*}{OLMo-3-7B-Think} & \multirow{3}{*}{1.0}
& GPTQ      & 7 & 4.28 \\
& & AWQ       & 7 & 2.04 \\
& & AutoRound & 5 & -0.19 \\
\hline \hline

\multirow{3}{*}{DeepSeek-R1-Distill-Qwen-1.5B} & \multirow{3}{*}{2.0}
& GPTQ      & 7 & 8.90 \\
& & AWQ       & 3 & 3.92 \\
& & AutoRound & 4 & 2.51 \\
\cline{1-5}

\multirow{3}{*}{DeepSeek-R1-Distill-Qwen-7B} & \multirow{3}{*}{2.0}
& GPTQ      & 6 & -5.40 \\
& & AWQ       & 5 & -3.83 \\
& & AutoRound & 4 & -6.75 \\
\cline{1-5}

\multirow{3}{*}{OLMo-3-7B-Think} & \multirow{3}{*}{2.0}
& GPTQ      & 8 & 4.28 \\
& & AWQ       & 7 & 2.04 \\
& & AutoRound & 7 & -0.19 \\

\bottomrule
\end{tabular}
\caption{
Sensitivity of tolerance-based TC voting to $\epsilon$. 
Methods within $\epsilon$ TC points of the best signed $\Delta$TC are considered as co-winners. DeepSeek-R1-Distill-Qwen-1.5B/ 7B and OLMo-3-7B-Think values are recorded for eight languages.
}
\label{tab:tc_tolerance_sensitivity}
\end{table*}

\section{Quantization Algorithms}
\label{sec:apdx_algo}

In this section, we briefly describe the quantization methods, evaluated in this study. 
All quantized models are produced using \texttt{llm-compressor} standard presets to ensure a fair and reproducible comparison across methods. 

\subsection{Integer W4A16 Quantization}

\paragraph{GPTQ.}
GPTQ~\citep{frantar2022gptq} is a post-training quantization method that reduces reconstruction error between full precision and quantized layer outputs of the model. 
Given input activation $\mathbf{X}$ and weight $\mathbf{W}$, GPTQ tries to calculate quantized weight matrix $\hat{\mathbf{W}}$ by solving:
\[
\arg\min_{\hat{\mathbf{W}}} 
\|\hat{\mathbf{W}}\mathbf{X} - \mathbf{W}\mathbf{X}\|_F .
\]
It uses approximate second-order information to reduce quantization error. 
In this study, GPTQ is evaluated using the \texttt{llm-compressor} W4A16 preset with group size $g{=}128$. 

\paragraph{AWQ.}
AWQ~\citep{lin2023awq} is an activation-aware post-training quantization method that protects salient weight channels by applying channel-wise scale factor before applying quantization. 
Given activations $\mathbf{X}$, weights $\mathbf{W}$, and scaling vector $\mathbf{s}$, AWQ rewrites the linear operation as:
\[
\mathbf{Y} = (\mathbf{X}\mathbf{s}^{-1})(\mathbf{s}\mathbf{W}^{\top}),
\]
where $\mathbf{s}$ is chosen to reduce quantization error due to activation outliers.  
We evaluate AWQ using \texttt{llm-compressor} W4A16 preset with group size $g{=}128$. 
Like GPTQ, we use calibration with 128 samples from the Pile~\citep{gao2020pile}.

\paragraph{AutoRound.}
AutoRound~\citep{cheng2024optimizeweightroundingsigned} is a post-training quantization method that uses signed gradient descent to optimize the weight rounding and clipping ranges values. 
Instead of relying on nearest rounding, it tries to learn the rounding decisions that reduce layer-wise reconstruction error with no extra inference time overhead. 
In our paper, we evaluate AutoRound using \texttt{llm-compressor} W4A16 preset with group size $g{=}128$.

\section{Prompting Strategy.}
\label{append:prompting_method}
For each problem, the model is instructed to reason in the target language and the target language is again used to place the final-answer inside a boxed expression. These thinking and answer formatting phrases for given target language, are also provided by PolyMath ~\citep{wang2025polymath}. The prompting design is given below

\begin{quote}
\small
\texttt{\{target\_language\_question\}} \\
\texttt{\{target\_language\_reasoning\} \{target\_language\_answer\_formatting\}}
\end{quote}

For English, the prompt becomes:

\begin{quote}
\small
\texttt{\{target\_language\_question\}} \\
\texttt{Use English to think and answer. \\ Note: Please put the final answer in \(\backslash\)boxed\{\}.}
\end{quote}

 
During post-processing, we split the model response into corresponding thinking and answer parts using the \texttt{</think>} delimiter if it is available. This way, we record raw response and corresponding reasoning and answer parts for the evaluations.

\section{Additional Experiments}
\label{append:additional-experiments}

\subsection{Tolerance-Based Voting}
\label{append:tolerance-based-voting}

We repeat tolerance-based voting with $\epsilon \in \{0.5,1.0,2.0\}$. 
For each model-language pair, the method with the highest signed $\Delta$TC is selected as the reference, and all methods within $\epsilon$ TC points are counted as co-winners. 
We report co-winner counts and average signed $\Delta$TC in Table~\ref{tab:tc_tolerance_sensitivity}.

\subsection{Breakdown-Threshold Sensitivity}
\label{sec:breakdown-threshold-sensitivity}

Table~\ref{tab:breakdown-threshold-sensitivity} reports breakdown counts
under $\Delta_{\text{break}}\in\{30,50,70\}$ percentage points.
As expected, the number of detected breakdowns generally decreases as the
threshold becomes stricter. However, the main conclusion remains same:
DeepSeek reasoning models exhibit breakdowns in both TC and AC, OLMo-3-7B-Think
and Phi-4-mini-reasoning primarily exhibit AC breakdowns, and
Qwen3-30B-A3B-Thinking-2507 and the evaluated non-reasoning models remain
comparatively stable.

\begin{table*}[t]
\centering
\small
\setlength{\tabcolsep}{5pt}
\begin{tabular}{lcccccc}
\toprule
& \multicolumn{2}{c}{$\Delta_{\text{break}}=30$}
& \multicolumn{2}{c}{$\Delta_{\text{break}}=50$}
& \multicolumn{2}{c}{$\Delta_{\text{break}}=70$} \\
\cmidrule(lr){2-3}
\cmidrule(lr){4-5}
\cmidrule(lr){6-7}
Model & TC & AC & TC & AC & TC & AC \\
\midrule
DeepSeek-R1-Distill-Qwen-1.5B
& 5/8 & 5/8 & 3/8 & 4/8 & 2/8 & 1/8 \\

DeepSeek-R1-Distill-Qwen-7B
& 5/8 & 5/8 & 5/8 & 5/8 & 4/8 & 5/8 \\

OLMo-3-7B-Think
& 0/8 & 6/8 & 0/8 & 4/8 & 0/8 & 2/8 \\

Phi-4-mini-reasoning
& 0/8 & 4/8 & 0/8 & 3/8 & 0/8 & 0/8 \\

Qwen3-30B-A3B-Thinking-2507
& 0/8 & 0/8 & 0/8 & 0/8 & 0/8 & 0/8 \\
\midrule
Gemma-3-12B-IT
& -- & 0/8 & -- & 0/8 & -- & 0/8 \\

Phi-4-mini-instruct
& -- & 0/8 & -- & 0/8 & -- & 0/8 \\

Qwen3-30B-A3B-Instruct-2507
& -- & 0/8 & -- & 0/8 & -- & 0/8 \\
\bottomrule
\end{tabular}
\caption{
Sensitivity of language-consistency breakdown counts to
$\Delta_{\text{break}} \in \{30,50,70\}$ percentage points.
Each value reports the number of languages, out of eight, exhibiting at
least one breakdown between consecutive difficulty levels.
}
\label{tab:breakdown-threshold-sensitivity}
\end{table*}

\begin{table*}[t]
\centering
\small
\setlength{\tabcolsep}{4.5pt}
\begin{tabular}{lccc}
\toprule
\textbf{Model} &
\textbf{Avg. Acc.} &
\textbf{TC below $\tau_{\text{cons}}$ @ Top} &
\textbf{AC below $\tau_{\text{cons}}$ @ Top} \\
\midrule

DeepSeek-R1-Distill-Qwen-7B
& 35.3 & 6/8 & 6/8 \\

Phi-4-mini-reasoning
& 32.4 & 6/8 & 7/8 \\

\bottomrule
\end{tabular}

\caption{
Task accuracy and language-consistency behavior under preferred-language reasoning prompting. Avg. Acc. denotes average PolyMath accuracy across eight languages. TC below $\tau_{\text{cons}}$ @ Top and AC below $\tau_{\text{cons}}$ @ Top count languages with thinking- and answer-language consistency below $\tau_{\text{cons}}=90\%$ at Top difficulty. The threshold $\tau_{\text{cons}}$ is an analysis hyperparameter.
}
\label{tab:preferred-language-prompt-control}
\end{table*}



 

\section{Hardware and Software Details}
\label{append:hardware-software}

Our experimental pipeline consists of three stages. quantization, inference, and evaluation. 

For quantization, we use \texttt{llm-compressor} version 0.9.0.2 and \texttt{compressed-tensors} version 0.13.0. 
For inference, we use \texttt{vLLM} version 0.20.1 with \texttt{torch} version 2.11.0 and \texttt{transformers} version 5.7.0. 
Evaluation is performed using the official PolyMath scripts, which compute final-answer correctness as well as language consistency scores.

All experiments are run on NVIDIA H100 GPUs.

\section{AI Assistance}
\label{append:ai-assistance}
ChatGPT-5.5 was used for grammar refinement and latex coding assistance with table/ figure structure and debugging. The authors take full responsibility for the final published version.

\clearpage

\begin{figure*}[p]
\centering

\addtocounter{figure}{-1}

Qwen3-30B-A3B-Instruct-2507


\vspace{0.25cm}
\begin{tikzpicture}
\begin{axis}[
    hide axis,
    xmin=0, xmax=1, ymin=0, ymax=1,
    width=15cm,
    height=6cm,
    legend columns=3,
    legend style={
        draw=black!30,
        fill=white,
        font=\small,
        at={(0.5,0.5)},
        anchor=center,
        column sep=0.2cm,
    },
]
\addlegendimage{
    color=green!60!black,
    dotted,
    mark=*,
    thick,
    mark size=2pt
}
\addlegendentry{Accuracy}


\addlegendimage{
    color=orange!90!black,
    solid,
    mark=*,
    thick,
    mark size=2pt
}
\addlegendentry{Answer Consistency (AC)}
\end{axis}
\end{tikzpicture}

\vspace{0.4cm}

\begin{subfigure}[b]{0.24\textwidth}
\centering
\begin{tikzpicture}
\begin{axis}[
    width=\textwidth,
    height=0.9\textwidth,
    xlabel={Difficulty Level},
    ylabel={Score (\%)},
    xmin=0.5, xmax=4.5,
    ymin=-5, ymax=105,
    xtick={1,2,3,4},
    ytick={0,20,40,60,80,100},
    tick label style={font=\small},
    label style={font=\small},
    grid=major,
    grid style={line width=.1pt, draw=gray!30},
    major grid style={line width=.2pt, draw=gray!50},
]

\addplot[
    color=green!60!black,
    dotted,
    mark=*,
    thick,
    mark size=1.0pt
] coordinates {
    (1,97.6) (2,61.6) (3,51.2) (4,29.6)
};


\addplot[
    color=orange!90!black,
    solid,
    mark=*,
    thick,
    mark size=1.0pt
] coordinates {
    (1,99.2) (2,100.0) (3,100.0) (4,100.0)
};

\end{axis}
\end{tikzpicture}
\caption{English}
\label{fig:en-qwen3-30b-a3b-instruct-combined}
\end{subfigure}
\hfill
\begin{subfigure}[b]{0.24\textwidth}
\centering
\begin{tikzpicture}
\begin{axis}[
    width=\textwidth,
    height=0.9\textwidth,
    xlabel={Difficulty Level},
    xmin=0.5, xmax=4.5,
    ymin=-5, ymax=105,
    xtick={1,2,3,4},
    ytick={0,20,40,60,80,100},
    tick label style={font=\small},
    label style={font=\small},
    grid=major,
    grid style={line width=.1pt, draw=gray!30},
    major grid style={line width=.2pt, draw=gray!50},
]

\addplot[
    color=green!60!black,
    dotted,
    mark=*,
    thick,
    mark size=1.0pt
] coordinates {
    (1,90.4) (2,59.2) (3,42.4) (4,24.0)
};


\addplot[
    color=orange!90!black,
    solid,
    mark=*,
    thick,
    mark size=1.0pt
] coordinates {
    (1,100.0) (2,95.8) (3,95.0) (4,96.4)
};

\end{axis}
\end{tikzpicture}
\caption{Chinese}
\label{fig:zh-qwen3-30b-a3b-instruct-combined}
\end{subfigure}
\hfill
\begin{subfigure}[b]{0.24\textwidth}
\centering
\begin{tikzpicture}
\begin{axis}[
    width=\textwidth,
    height=0.9\textwidth,
    xlabel={Difficulty Level},
    xmin=0.5, xmax=4.5,
    ymin=-5, ymax=105,
    xtick={1,2,3,4},
    ytick={0,20,40,60,80,100},
    tick label style={font=\small},
    label style={font=\small},
    grid=major,
    grid style={line width=.1pt, draw=gray!30},
    major grid style={line width=.2pt, draw=gray!50},
]

\addplot[
    color=green!60!black,
    dotted,
    mark=*,
    thick,
    mark size=1.0pt
] coordinates {
    (1,89.6) (2,64.0) (3,46.4) (4,21.6)
};


\addplot[
    color=orange!90!black,
    solid,
    mark=*,
    thick,
    mark size=1.0pt
] coordinates {
    (1,100.0) (2,100.0) (3,100.0) (4,100.0)
};

\end{axis}
\end{tikzpicture}
\caption{Arabic}
\label{fig:ar-qwen3-30b-a3b-instruct-combined}
\end{subfigure}
\hfill
\begin{subfigure}[b]{0.24\textwidth}
\centering
\begin{tikzpicture}
\begin{axis}[
    width=\textwidth,
    height=0.9\textwidth,
    xlabel={Difficulty Level},
    xmin=0.5, xmax=4.5,
    ymin=-5, ymax=105,
    xtick={1,2,3,4},
    ytick={0,20,40,60,80,100},
    tick label style={font=\small},
    label style={font=\small},
    grid=major,
    grid style={line width=.1pt, draw=gray!30},
    major grid style={line width=.2pt, draw=gray!50},
]

\addplot[
    color=green!60!black,
    dotted,
    mark=*,
    thick,
    mark size=1.0pt
] coordinates {
    (1,87.2) (2,56.8) (3,35.2) (4,24.0)
};


\addplot[
    color=orange!90!black,
    solid,
    mark=*,
    thick,
    mark size=1.0pt
] coordinates {
    (1,100.0) (2,100.0) (3,100.0) (4,100.0)
};

\end{axis}
\end{tikzpicture}
\caption{Bengali}
\label{fig:bn-qwen3-30b-a3b-instruct-combined}
\end{subfigure}

\vspace{0.5cm}

\begin{subfigure}[b]{0.24\textwidth}
\centering
\begin{tikzpicture}
\begin{axis}[
    width=\textwidth,
    height=0.9\textwidth,
    xlabel={Difficulty Level},
    ylabel={Score (\%)},
    xmin=0.5, xmax=4.5,
    ymin=-5, ymax=105,
    xtick={1,2,3,4},
    ytick={0,20,40,60,80,100},
    tick label style={font=\small},
    label style={font=\small},
    grid=major,
    grid style={line width=.1pt, draw=gray!30},
    major grid style={line width=.2pt, draw=gray!50},
]

\addplot[
    color=green!60!black,
    dotted,
    mark=*,
    thick,
    mark size=1.0pt
] coordinates {
    (1,86.4) (2,67.2) (3,48.0) (4,26.4)
};


\addplot[
    color=orange!90!black,
    solid,
    mark=*,
    thick,
    mark size=1.0pt
] coordinates {
    (1,99.2) (2,100.0) (3,100.0) (4,100.0)
};

\end{axis}
\end{tikzpicture}
\caption{French}
\label{fig:fr-qwen3-30b-a3b-instruct-combined}
\end{subfigure}
\hfill
\begin{subfigure}[b]{0.24\textwidth}
\centering
\begin{tikzpicture}
\begin{axis}[
    width=\textwidth,
    height=0.9\textwidth,
    xlabel={Difficulty Level},
    xmin=0.5, xmax=4.5,
    ymin=-5, ymax=105,
    xtick={1,2,3,4},
    ytick={0,20,40,60,80,100},
    tick label style={font=\small},
    label style={font=\small},
    grid=major,
    grid style={line width=.1pt, draw=gray!30},
    major grid style={line width=.2pt, draw=gray!50},
]

\addplot[
    color=green!60!black,
    dotted,
    mark=*,
    thick,
    mark size=1.0pt
] coordinates {
    (1,92.0) (2,63.2) (3,44.8) (4,24.8)
};


\addplot[
    color=orange!90!black,
    solid,
    mark=*,
    thick,
    mark size=1.0pt
] coordinates {
    (1,98.4) (2,100.0) (3,100.0) (4,100.0)
};

\end{axis}
\end{tikzpicture}
\caption{Russian}
\label{fig:ru-qwen3-30b-a3b-instruct-combined}
\end{subfigure}
\hfill
\begin{subfigure}[b]{0.24\textwidth}
\centering
\begin{tikzpicture}
\begin{axis}[
    width=\textwidth,
    height=0.9\textwidth,
    xlabel={Difficulty Level},
    xmin=0.5, xmax=4.5,
    ymin=-5, ymax=105,
    xtick={1,2,3,4},
    ytick={0,20,40,60,80,100},
    tick label style={font=\small},
    label style={font=\small},
    grid=major,
    grid style={line width=.1pt, draw=gray!30},
    major grid style={line width=.2pt, draw=gray!50},
]

\addplot[
    color=green!60!black,
    dotted,
    mark=*,
    thick,
    mark size=1.0pt
] coordinates {
    (1,54.4) (2,33.6) (3,12.0) (4,4.8)
};


\addplot[
    color=orange!90!black,
    solid,
    mark=*,
    thick,
    mark size=1.0pt
] coordinates {
    (1,100.0) (2,98.4) (3,100.0) (4,100.0)
};

\end{axis}
\end{tikzpicture}
\caption{Swahili}
\label{fig:sw-qwen3-30b-a3b-instruct-combined}
\end{subfigure}
\hfill
\begin{subfigure}[b]{0.24\textwidth}
\centering
\begin{tikzpicture}
\begin{axis}[
    width=\textwidth,
    height=0.9\textwidth,
    xlabel={Difficulty Level},
    xmin=0.5, xmax=4.5,
    ymin=-5, ymax=105,
    xtick={1,2,3,4},
    ytick={0,20,40,60,80,100},
    tick label style={font=\small},
    label style={font=\small},
    grid=major,
    grid style={line width=.1pt, draw=gray!30},
    major grid style={line width=.2pt, draw=gray!50},
]

\addplot[
    color=green!60!black,
    dotted,
    mark=*,
    thick,
    mark size=1.0pt
] coordinates {
    (1,78.4) (2,51.2) (3,28.0) (4,16.8)
};


\addplot[
    color=orange!90!black,
    solid,
    mark=*,
    thick,
    mark size=1.0pt
] coordinates {
    (1,100.0) (2,100.0) (3,100.0) (4,100.0)
};

\end{axis}
\end{tikzpicture}
\caption{Telugu}
\label{fig:te-qwen3-30b-a3b-instruct-combined}
\end{subfigure}

\captionof{figure}{
Difficulty-wise final-answer accuracy, thinking-language consistency (TC),
and answer-language consistency (AC) of Qwen3-30B-A3B-Instruct-2507 across eight languages and four PolyMath difficulty levels: 1 = low, 2 = medium, 3 = high, and 4 = top.
The results show that answer-language
consistency is largely preserved across several languages.
}
\label{fig:bf16_accuracy_tc_ac_difficulty_qwen3_30b_a3b_instruct}


\vspace{0.4cm}

Qwen3-30B-A3B-Thinking-2507


\vspace{0.25cm}

\begin{tikzpicture}
\begin{axis}[
    hide axis,
    xmin=0, xmax=1, ymin=0, ymax=1,
    width=15cm,
    height=6cm,
    legend columns=3,
    legend style={
        draw=black!30,
        fill=white,
        font=\small,
        at={(0.5,0.5)},
        anchor=center,
        column sep=0.2cm,
    },
]
\addlegendimage{
    color=green!60!black,
    dotted,
    mark=*,
    thick,
    mark size=2pt
}
\addlegendentry{Accuracy}

\addlegendimage{
    color=orange!90!black,
    solid,
    mark=*,
    thick,
    mark size=2pt
}
\addlegendentry{Answer Consistency (AC)}

\addlegendimage{
    color=red!85!black,
    solid,
    mark=*,
    thick,
    mark size=2pt
}
\addlegendentry{Thinking Consistency (TC)}

\end{axis}
\end{tikzpicture}

\vspace{0.4cm}

\begin{subfigure}[b]{0.24\textwidth}
\centering
\begin{tikzpicture}
\begin{axis}[
    width=\textwidth,
    height=0.9\textwidth,
    xlabel={Difficulty Level},
    ylabel={Score (\%)},
    xmin=0.5, xmax=4.5,
    ymin=-5, ymax=105,
    xtick={1,2,3,4},
    ytick={0,20,40,60,80,100},
    tick label style={font=\small},
    label style={font=\small},
    grid=major,
    grid style={line width=.1pt, draw=gray!30},
    major grid style={line width=.2pt, draw=gray!50},
]

\addplot[
    color=green!60!black,
    dotted,
    mark=*,
    thick,
    mark size=1.0pt
] coordinates {
    (1,98.4) (2,74.4) (3,68.8) (4,38.4)
};

\addplot[
    color=red!85!black,
    solid,
    mark=*,
    thick,
    mark size=1.0pt
] coordinates {
    (1,100.0) (2,100.0) (3,100.0) (4,100.0)
};

\addplot[
    color=orange!90!black,
    solid,
    mark=*,
    thick,
    mark size=1.0pt
] coordinates {
    (1,99.2) (2,100.0) (3,100.0) (4,100.0)
};

\end{axis}
\end{tikzpicture}
\caption{English}
\label{fig:en-qwen3-30b-a3b-thinking-combined}
\end{subfigure}
\hfill
\begin{subfigure}[b]{0.24\textwidth}
\centering
\begin{tikzpicture}
\begin{axis}[
    width=\textwidth,
    height=0.9\textwidth,
    xlabel={Difficulty Level},
    xmin=0.5, xmax=4.5,
    ymin=-5, ymax=105,
    xtick={1,2,3,4},
    ytick={0,20,40,60,80,100},
    tick label style={font=\small},
    label style={font=\small},
    grid=major,
    grid style={line width=.1pt, draw=gray!30},
    major grid style={line width=.2pt, draw=gray!50},
]

\addplot[
    color=green!60!black,
    dotted,
    mark=*,
    thick,
    mark size=1.0pt
] coordinates {
    (1,92.0) (2,70.4) (3,58.4) (4,31.2)
};

\addplot[
    color=red!85!black,
    solid,
    mark=*,
    thick,
    mark size=1.0pt
] coordinates {
    (1,100.0) (2,96.8) (3,92.4) (4,93.2)
};

\addplot[
    color=orange!90!black,
    solid,
    mark=*,
    thick,
    mark size=1.0pt
] coordinates {
    (1,100.0) (2,100.0) (3,100.0) (4,100.0)
};

\end{axis}
\end{tikzpicture}
\caption{Chinese}
\label{fig:zh-qwen3-30b-a3b-thinking-combined}
\end{subfigure}
\hfill
\begin{subfigure}[b]{0.24\textwidth}
\centering
\begin{tikzpicture}
\begin{axis}[
    width=\textwidth,
    height=0.9\textwidth,
    xlabel={Difficulty Level},
    xmin=0.5, xmax=4.5,
    ymin=-5, ymax=105,
    xtick={1,2,3,4},
    ytick={0,20,40,60,80,100},
    tick label style={font=\small},
    label style={font=\small},
    grid=major,
    grid style={line width=.1pt, draw=gray!30},
    major grid style={line width=.2pt, draw=gray!50},
]

\addplot[
    color=green!60!black,
    dotted,
    mark=*,
    thick,
    mark size=1.0pt
] coordinates {
    (1,93.6) (2,70.4) (3,64.0) (4,33.6)
};

\addplot[
    color=red!85!black,
    solid,
    mark=*,
    thick,
    mark size=1.0pt
] coordinates {
    (1,0.0) (2,0.0) (3,0.0) (4,0.0)
};

\addplot[
    color=orange!90!black,
    solid,
    mark=*,
    thick,
    mark size=1.0pt
] coordinates {
    (1,100.0) (2,100.0) (3,100.0) (4,100.0)
};

\end{axis}
\end{tikzpicture}
\caption{Arabic}
\label{fig:ar-qwen3-30b-a3b-thinking-combined}
\end{subfigure}
\hfill
\begin{subfigure}[b]{0.24\textwidth}
\centering
\begin{tikzpicture}
\begin{axis}[
    width=\textwidth,
    height=0.9\textwidth,
    xlabel={Difficulty Level},
    xmin=0.5, xmax=4.5,
    ymin=-5, ymax=105,
    xtick={1,2,3,4},
    ytick={0,20,40,60,80,100},
    tick label style={font=\small},
    label style={font=\small},
    grid=major,
    grid style={line width=.1pt, draw=gray!30},
    major grid style={line width=.2pt, draw=gray!50},
]

\addplot[
    color=green!60!black,
    dotted,
    mark=*,
    thick,
    mark size=1.0pt
] coordinates {
    (1,92.8) (2,71.2) (3,60.0) (4,27.2)
};

\addplot[
    color=red!85!black,
    solid,
    mark=*,
    thick,
    mark size=1.0pt
] coordinates {
    (1,0.0) (2,0.0) (3,0.0) (4,0.0)
};

\addplot[
    color=orange!90!black,
    solid,
    mark=*,
    thick,
    mark size=1.0pt
] coordinates {
    (1,100.0) (2,100.0) (3,100.0) (4,99.2)
};

\end{axis}
\end{tikzpicture}
\caption{Bengali}
\label{fig:bn-qwen3-30b-a3b-thinking-combined}
\end{subfigure}

\vspace{0.5cm}

\begin{subfigure}[b]{0.24\textwidth}
\centering
\begin{tikzpicture}
\begin{axis}[
    width=\textwidth,
    height=0.9\textwidth,
    xlabel={Difficulty Level},
    ylabel={Score (\%)},
    xmin=0.5, xmax=4.5,
    ymin=-5, ymax=105,
    xtick={1,2,3,4},
    ytick={0,20,40,60,80,100},
    tick label style={font=\small},
    label style={font=\small},
    grid=major,
    grid style={line width=.1pt, draw=gray!30},
    major grid style={line width=.2pt, draw=gray!50},
]

\addplot[
    color=green!60!black,
    dotted,
    mark=*,
    thick,
    mark size=1.0pt
] coordinates {
    (1,88.8) (2,72.0) (3,62.4) (4,30.4)
};

\addplot[
    color=red!85!black,
    solid,
    mark=*,
    thick,
    mark size=1.0pt
] coordinates {
    (1,0.0) (2,0.0) (3,0.0) (4,0.0)
};

\addplot[
    color=orange!90!black,
    solid,
    mark=*,
    thick,
    mark size=1.0pt
] coordinates {
    (1,97.6) (2,100.0) (3,100.0) (4,100.0)
};

\end{axis}
\end{tikzpicture}
\caption{French}
\label{fig:fr-qwen3-30b-a3b-thinking-combined}
\end{subfigure}
\hfill
\begin{subfigure}[b]{0.24\textwidth}
\centering
\begin{tikzpicture}
\begin{axis}[
    width=\textwidth,
    height=0.9\textwidth,
    xlabel={Difficulty Level},
    xmin=0.5, xmax=4.5,
    ymin=-5, ymax=105,
    xtick={1,2,3,4},
    ytick={0,20,40,60,80,100},
    tick label style={font=\small},
    label style={font=\small},
    grid=major,
    grid style={line width=.1pt, draw=gray!30},
    major grid style={line width=.2pt, draw=gray!50},
]

\addplot[
    color=green!60!black,
    dotted,
    mark=*,
    thick,
    mark size=1.0pt
] coordinates {
    (1,92.8) (2,71.2) (3,59.2) (4,26.4)
};

\addplot[
    color=red!85!black,
    solid,
    mark=*,
    thick,
    mark size=1.0pt
] coordinates {
    (1,100.0) (2,100.0) (3,100.0) (4,99.2)
};

\addplot[
    color=orange!90!black,
    solid,
    mark=*,
    thick,
    mark size=1.0pt
] coordinates {
    (1,100.0) (2,100.0) (3,100.0) (4,100.0)
};

\end{axis}
\end{tikzpicture}
\caption{Russian}
\label{fig:ru-qwen3-30b-a3b-thinking-combined}
\end{subfigure}
\hfill
\begin{subfigure}[b]{0.24\textwidth}
\centering
\begin{tikzpicture}
\begin{axis}[
    width=\textwidth,
    height=0.9\textwidth,
    xlabel={Difficulty Level},
    xmin=0.5, xmax=4.5,
    ymin=-5, ymax=105,
    xtick={1,2,3,4},
    ytick={0,20,40,60,80,100},
    tick label style={font=\small},
    label style={font=\small},
    grid=major,
    grid style={line width=.1pt, draw=gray!30},
    major grid style={line width=.2pt, draw=gray!50},
]

\addplot[
    color=green!60!black,
    dotted,
    mark=*,
    thick,
    mark size=1.0pt
] coordinates {
    (1,72.8) (2,59.2) (3,51.2) (4,26.4)
};

\addplot[
    color=red!85!black,
    solid,
    mark=*,
    thick,
    mark size=1.0pt
] coordinates {
    (1,0.0) (2,0.0) (3,0.0) (4,0.8)
};

\addplot[
    color=orange!90!black,
    solid,
    mark=*,
    thick,
    mark size=1.0pt
] coordinates {
    (1,100.0) (2,97.6) (3,96.8) (4,96.8)
};

\end{axis}
\end{tikzpicture}
\caption{Swahili}
\label{fig:sw-qwen3-30b-a3b-thinking-combined}
\end{subfigure}
\hfill
\begin{subfigure}[b]{0.24\textwidth}
\centering
\begin{tikzpicture}
\begin{axis}[
    width=\textwidth,
    height=0.9\textwidth,
    xlabel={Difficulty Level},
    xmin=0.5, xmax=4.5,
    ymin=-5, ymax=105,
    xtick={1,2,3,4},
    ytick={0,20,40,60,80,100},
    tick label style={font=\small},
    label style={font=\small},
    grid=major,
    grid style={line width=.1pt, draw=gray!30},
    major grid style={line width=.2pt, draw=gray!50},
]

\addplot[
    color=green!60!black,
    dotted,
    mark=*,
    thick,
    mark size=1.0pt
] coordinates {
    (1,87.2) (2,67.2) (3,55.2) (4,26.4)
};

\addplot[
    color=red!85!black,
    solid,
    mark=*,
    thick,
    mark size=1.0pt
] coordinates {
    (1,0.0) (2,0.0) (3,0.0) (4,0.0)
};

\addplot[
    color=orange!90!black,
    solid,
    mark=*,
    thick,
    mark size=1.0pt
] coordinates {
    (1,100.0) (2,100.0) (3,95.2) (4,97.6)
};

\end{axis}
\end{tikzpicture}
\caption{Telugu}
\label{fig:te-qwen3-30b-a3b-thinking-combined}
\end{subfigure}

\captionof{figure}{
Difficulty-wise final-answer accuracy, thinking-language consistency (TC),
and answer-language consistency (AC) of Qwen3-30B-A3B-Thinking-2507 across eight languages and four PolyMath difficulty levels: 1 = low, 2 = medium, 3 = high, and 4 = top.
The results show that answer-language consistency is largely preserved, while the model often reasons in English or Chinese for Arabic, Bengali, French, Swahili, and Telugu.
}
\label{fig:bf16_accuracy_tc_ac_difficulty_qwen3_30b_a3b_thinking}


\end{figure*}

\clearpage

\begin{figure*}[p]
\centering

\addtocounter{figure}{-1}

DeepSeek-R1-Distill-Qwen-1.5B


\vspace{0.25cm}
\begin{tikzpicture}
\begin{axis}[
    hide axis,
    xmin=0, xmax=1, ymin=0, ymax=1,
    width=15cm,
    height=6cm,
    legend columns=3,
    legend style={
        draw=black!30,
        fill=white,
        font=\small,
        at={(0.5,0.5)},
        anchor=center,
        column sep=0.2cm,
    },
]
\addlegendimage{
    color=green!60!black,
    dotted,
    mark=*,
    thick,
    mark size=2pt
}
\addlegendentry{Accuracy}

\addlegendimage{
    color=orange!90!black,
    solid,
    mark=*,
    thick,
    mark size=2pt
}
\addlegendentry{Answer Consistency (AC)}

\addlegendimage{
    color=red!85!black,
    solid,
    mark=*,
    thick,
    mark size=2pt
}
\addlegendentry{Thinking Consistency (TC)}

\end{axis}
\end{tikzpicture}

\vspace{0.4cm}

\begin{subfigure}[b]{0.24\textwidth}
\centering
\begin{tikzpicture}
\begin{axis}[
    width=\textwidth,
    height=0.9\textwidth,
    xlabel={Difficulty Level},
    ylabel={Score (\%)},
    xmin=0.5, xmax=4.5,
    ymin=-5, ymax=105,
    xtick={1,2,3,4},
    ytick={0,20,40,60,80,100},
    tick label style={font=\small},
    label style={font=\small},
    grid=major,
    grid style={line width=.1pt, draw=gray!30},
    major grid style={line width=.2pt, draw=gray!50},
]

\addplot[
    color=green!60!black,
    dotted,
    mark=*,
    thick,
    mark size=1.0pt
] coordinates {
    (1,73.3) (2,41.6) (3,18.7) (4,9.1)
};

\addplot[
    color=red!85!black,
    solid,
    mark=*,
    thick,
    mark size=1.0pt
] coordinates {
    (1,100.0) (2,99.7) (3,99.7) (4,99.7)
};

\addplot[
    color=orange!90!black,
    solid,
    mark=*,
    thick,
    mark size=1.0pt
] coordinates {
    (1,98.9) (2,99.7) (3,100.0) (4,100.0)
};

\end{axis}
\end{tikzpicture}
\caption{English}
\label{fig:en-deepseek-r1-distill-qwen-1p5b-combined}
\end{subfigure}
\hfill
\begin{subfigure}[b]{0.24\textwidth}
\centering
\begin{tikzpicture}
\begin{axis}[
    width=\textwidth,
    height=0.9\textwidth,
    xlabel={Difficulty Level},
    xmin=0.5, xmax=4.5,
    ymin=-5, ymax=105,
    xtick={1,2,3,4},
    ytick={0,20,40,60,80,100},
    tick label style={font=\small},
    label style={font=\small},
    grid=major,
    grid style={line width=.1pt, draw=gray!30},
    major grid style={line width=.2pt, draw=gray!50},
]

\addplot[
    color=green!60!black,
    dotted,
    mark=*,
    thick,
    mark size=1.0pt
] coordinates {
    (1,63.2) (2,37.3) (3,13.9) (4,5.6)
};

\addplot[
    color=red!85!black,
    solid,
    mark=*,
    thick,
    mark size=1.0pt
] coordinates {
    (1,94.8) (2,94.3) (3,95.6) (4,93.9)
};

\addplot[
    color=orange!90!black,
    solid,
    mark=*,
    thick,
    mark size=1.0pt
] coordinates {
    (1,92.1) (2,96.4) (3,99.1) (4,97.5)
};

\end{axis}
\end{tikzpicture}
\caption{Chinese}
\label{fig:zh-deepseek-r1-distill-qwen-1p5b-combined}
\end{subfigure}
\hfill
\begin{subfigure}[b]{0.24\textwidth}
\centering
\begin{tikzpicture}
\begin{axis}[
    width=\textwidth,
    height=0.9\textwidth,
    xlabel={Difficulty Level},
    xmin=0.5, xmax=4.5,
    ymin=-5, ymax=105,
    xtick={1,2,3,4},
    ytick={0,20,40,60,80,100},
    tick label style={font=\small},
    label style={font=\small},
    grid=major,
    grid style={line width=.1pt, draw=gray!30},
    major grid style={line width=.2pt, draw=gray!50},
]

\addplot[
    color=green!60!black,
    dotted,
    mark=*,
    thick,
    mark size=1.0pt
] coordinates {
    (1,16.8) (2,34.7) (3,14.7) (4,8.5)
};

\addplot[
    color=red!85!black,
    solid,
    mark=*,
    thick,
    mark size=1.0pt
] coordinates {
    (1,78.4) (2,7.2) (3,12.3) (4,14.1)
};

\addplot[
    color=orange!90!black,
    solid,
    mark=*,
    thick,
    mark size=1.0pt
] coordinates {
    (1,68.5) (2,0.8) (3,2.1) (4,3.5)
};

\end{axis}
\end{tikzpicture}
\caption{Arabic}
\label{fig:ar-deepseek-r1-distill-qwen-1p5b-combined}
\end{subfigure}
\hfill
\begin{subfigure}[b]{0.24\textwidth}
\centering
\begin{tikzpicture}
\begin{axis}[
    width=\textwidth,
    height=0.9\textwidth,
    xlabel={Difficulty Level},
    xmin=0.5, xmax=4.5,
    ymin=-5, ymax=105,
    xtick={1,2,3,4},
    ytick={0,20,40,60,80,100},
    tick label style={font=\small},
    label style={font=\small},
    grid=major,
    grid style={line width=.1pt, draw=gray!30},
    major grid style={line width=.2pt, draw=gray!50},
]

\addplot[
    color=green!60!black,
    dotted,
    mark=*,
    thick,
    mark size=1.0pt
] coordinates {
    (1,13.9) (2,29.6) (3,11.7) (4,5.9)
};

\addplot[
    color=red!85!black,
    solid,
    mark=*,
    thick,
    mark size=1.0pt
] coordinates {
    (1,11.2) (2,2.4) (3,7.5) (4,9.3)
};

\addplot[
    color=orange!90!black,
    solid,
    mark=*,
    thick,
    mark size=1.0pt
] coordinates {
    (1,8.0) (2,0.5) (3,0.3) (4,0.3)
};

\end{axis}
\end{tikzpicture}
\caption{Bengali}
\label{fig:bn-deepseek-r1-distill-qwen-1p5b-combined}
\end{subfigure}

\vspace{0.5cm}

\begin{subfigure}[b]{0.24\textwidth}
\centering
\begin{tikzpicture}
\begin{axis}[
    width=\textwidth,
    height=0.9\textwidth,
    xlabel={Difficulty Level},
    ylabel={Score (\%)},
    xmin=0.5, xmax=4.5,
    ymin=-5, ymax=105,
    xtick={1,2,3,4},
    ytick={0,20,40,60,80,100},
    tick label style={font=\small},
    label style={font=\small},
    grid=major,
    grid style={line width=.1pt, draw=gray!30},
    major grid style={line width=.2pt, draw=gray!50},
]

\addplot[
    color=green!60!black,
    dotted,
    mark=*,
    thick,
    mark size=1.0pt
] coordinates {
    (1,40.3) (2,35.7) (3,13.3) (4,7.7)
};

\addplot[
    color=red!85!black,
    solid,
    mark=*,
    thick,
    mark size=1.0pt
] coordinates {
    (1,99.2) (2,28.8) (3,30.7) (4,24.5)
};

\addplot[
    color=orange!90!black,
    solid,
    mark=*,
    thick,
    mark size=1.0pt
] coordinates {
    (1,98.1) (2,24.0) (3,21.1) (4,15.5)
};

\end{axis}
\end{tikzpicture}
\caption{French}
\label{fig:fr-deepseek-r1-distill-qwen-1p5b-combined}
\end{subfigure}
\hfill
\begin{subfigure}[b]{0.24\textwidth}
\centering
\begin{tikzpicture}
\begin{axis}[
    width=\textwidth,
    height=0.9\textwidth,
    xlabel={Difficulty Level},
    xmin=0.5, xmax=4.5,
    ymin=-5, ymax=105,
    xtick={1,2,3,4},
    ytick={0,20,40,60,80,100},
    tick label style={font=\small},
    label style={font=\small},
    grid=major,
    grid style={line width=.1pt, draw=gray!30},
    major grid style={line width=.2pt, draw=gray!50},
]

\addplot[
    color=green!60!black,
    dotted,
    mark=*,
    thick,
    mark size=1.0pt
] coordinates {
    (1,44.8) (2,27.2) (3,10.7) (4,3.5)
};

\addplot[
    color=red!85!black,
    solid,
    mark=*,
    thick,
    mark size=1.0pt
] coordinates {
    (1,93.3) (2,49.9) (3,60.8) (4,61.1)
};

\addplot[
    color=orange!90!black,
    solid,
    mark=*,
    thick,
    mark size=1.0pt
] coordinates {
    (1,85.3) (2,30.4) (3,29.9) (4,30.9)
};

\end{axis}
\end{tikzpicture}
\caption{Russian}
\label{fig:ru-deepseek-r1-distill-qwen-1p5b-combined}
\end{subfigure}
\hfill
\begin{subfigure}[b]{0.24\textwidth}
\centering
\begin{tikzpicture}
\begin{axis}[
    width=\textwidth,
    height=0.9\textwidth,
    xlabel={Difficulty Level},
    xmin=0.5, xmax=4.5,
    ymin=-5, ymax=105,
    xtick={1,2,3,4},
    ytick={0,20,40,60,80,100},
    tick label style={font=\small},
    label style={font=\small},
    grid=major,
    grid style={line width=.1pt, draw=gray!30},
    major grid style={line width=.2pt, draw=gray!50},
]

\addplot[
    color=green!60!black,
    dotted,
    mark=*,
    thick,
    mark size=1.0pt
] coordinates {
    (1,2.9) (2,18.1) (3,7.2) (4,2.1)
};

\addplot[
    color=red!85!black,
    solid,
    mark=*,
    thick,
    mark size=1.0pt
] coordinates {
    (1,41.9) (2,8.3) (3,15.7) (4,13.9)
};

\addplot[
    color=orange!90!black,
    solid,
    mark=*,
    thick,
    mark size=1.0pt
] coordinates {
    (1,48.8) (2,1.3) (3,5.3) (4,6.7)
};

\end{axis}
\end{tikzpicture}
\caption{Swahili}
\label{fig:sw-deepseek-r1-distill-qwen-1p5b-combined}
\end{subfigure}
\hfill
\begin{subfigure}[b]{0.24\textwidth}
\centering
\begin{tikzpicture}
\begin{axis}[
    width=\textwidth,
    height=0.9\textwidth,
    xlabel={Difficulty Level},
    xmin=0.5, xmax=4.5,
    ymin=-5, ymax=105,
    xtick={1,2,3,4},
    ytick={0,20,40,60,80,100},
    tick label style={font=\small},
    label style={font=\small},
    grid=major,
    grid style={line width=.1pt, draw=gray!30},
    major grid style={line width=.2pt, draw=gray!50},
]

\addplot[
    color=green!60!black,
    dotted,
    mark=*,
    thick,
    mark size=1.0pt
] coordinates {
    (1,1.1) (2,22.9) (3,10.1) (4,3.7)
};

\addplot[
    color=red!85!black,
    solid,
    mark=*,
    thick,
    mark size=1.0pt
] coordinates {
    (1,66.1) (2,12.5) (3,12.0) (4,10.9)
};

\addplot[
    color=orange!90!black,
    solid,
    mark=*,
    thick,
    mark size=1.0pt
] coordinates {
    (1,61.1) (2,6.4) (3,4.3) (4,2.9)
};

\end{axis}
\end{tikzpicture}
\caption{Telugu}
\label{fig:te-deepseek-r1-distill-qwen-1p5b-combined}
\end{subfigure}

\captionof{figure}{
Difficulty-wise final-answer accuracy, thinking-language consistency (TC),
and answer-language consistency (AC) of DeepSeek-R1-Distill-Qwen-1.5B across eight languages and four PolyMath difficulty levels: 1 = low, 2 = medium, 3 = high, and 4 = top.
The results show that language consistency is preserved for English and Chinese but breaks down for several other languages.
}
\label{fig:bf16_accuracy_tc_ac_difficulty_deepseek_r1_distill_qwen_1p5b}


\vspace{0.4cm}

Gemma-3-12B-IT


\vspace{0.25cm}

\begin{tikzpicture}
\begin{axis}[
    hide axis,
    xmin=0, xmax=1, ymin=0, ymax=1,
    width=15cm,
    height=6cm,
    legend columns=3,
    legend style={
        draw=black!30,
        fill=white,
        font=\small,
        at={(0.5,0.5)},
        anchor=center,
        column sep=0.2cm,
    },
]
\addlegendimage{
    color=green!60!black,
    dotted,
    mark=*,
    thick,
    mark size=2pt
}
\addlegendentry{Accuracy}


\addlegendimage{
    color=orange!90!black,
    solid,
    mark=*,
    thick,
    mark size=2pt
}
\addlegendentry{Answer Consistency (AC)}
\end{axis}
\end{tikzpicture}

\vspace{0.4cm}

\begin{subfigure}[b]{0.24\textwidth}
\centering
\begin{tikzpicture}
\begin{axis}[
    width=\textwidth,
    height=0.9\textwidth,
    xlabel={Difficulty Level},
    ylabel={Score (\%)},
    xmin=0.5, xmax=4.5,
    ymin=-5, ymax=105,
    xtick={1,2,3,4},
    ytick={0,20,40,60,80,100},
    tick label style={font=\small},
    label style={font=\small},
    grid=major,
    grid style={line width=.1pt, draw=gray!30},
    major grid style={line width=.2pt, draw=gray!50},
]

\addplot[
    color=green!60!black,
    dotted,
    mark=*,
    thick,
    mark size=1.0pt
] coordinates {
    (1,91.7) (2,35.2) (3,16.0) (4,7.5)
};


\addplot[
    color=orange!90!black,
    solid,
    mark=*,
    thick,
    mark size=1.0pt
] coordinates {
    (1,98.1) (2,99.7) (3,99.7) (4,99.2)
};

\end{axis}
\end{tikzpicture}
\caption{English}
\label{fig:en-gemma3-12b-it-combined}
\end{subfigure}
\hfill
\begin{subfigure}[b]{0.24\textwidth}
\centering
\begin{tikzpicture}
\begin{axis}[
    width=\textwidth,
    height=0.9\textwidth,
    xlabel={Difficulty Level},
    xmin=0.5, xmax=4.5,
    ymin=-5, ymax=105,
    xtick={1,2,3,4},
    ytick={0,20,40,60,80,100},
    tick label style={font=\small},
    label style={font=\small},
    grid=major,
    grid style={line width=.1pt, draw=gray!30},
    major grid style={line width=.2pt, draw=gray!50},
]

\addplot[
    color=green!60!black,
    dotted,
    mark=*,
    thick,
    mark size=1.0pt
] coordinates {
    (1,55.5) (2,32.0) (3,13.9) (4,8.8)
};


\addplot[
    color=orange!90!black,
    solid,
    mark=*,
    thick,
    mark size=1.0pt
] coordinates {
    (1,100) (2,99.5) (3,98.4) (4,99.2)
};

\end{axis}
\end{tikzpicture}
\caption{Chinese}
\label{fig:zh-gemma3-12b-it-combined}
\end{subfigure}
\hfill
\begin{subfigure}[b]{0.24\textwidth}
\centering
\begin{tikzpicture}
\begin{axis}[
    width=\textwidth,
    height=0.9\textwidth,
    xlabel={Difficulty Level},
    xmin=0.5, xmax=4.5,
    ymin=-5, ymax=105,
    xtick={1,2,3,4},
    ytick={0,20,40,60,80,100},
    tick label style={font=\small},
    label style={font=\small},
    grid=major,
    grid style={line width=.1pt, draw=gray!30},
    major grid style={line width=.2pt, draw=gray!50},
]

\addplot[
    color=green!60!black,
    dotted,
    mark=*,
    thick,
    mark size=1.0pt
] coordinates {
    (1,85.9) (2,28.3) (3,11.2) (4,5.9)
};


\addplot[
    color=orange!90!black,
    solid,
    mark=*,
    thick,
    mark size=1.0pt
] coordinates {
    (1,100.0) (2,98.1) (3,98.1) (4,98.7)
};

\end{axis}
\end{tikzpicture}
\caption{Arabic}
\label{fig:ar-gemma3-12b-it-combined}
\end{subfigure}
\hfill
\begin{subfigure}[b]{0.24\textwidth}
\centering
\begin{tikzpicture}
\begin{axis}[
    width=\textwidth,
    height=0.9\textwidth,
    xlabel={Difficulty Level},
    xmin=0.5, xmax=4.5,
    ymin=-5, ymax=105,
    xtick={1,2,3,4},
    ytick={0,20,40,60,80,100},
    tick label style={font=\small},
    label style={font=\small},
    grid=major,
    grid style={line width=.1pt, draw=gray!30},
    major grid style={line width=.2pt, draw=gray!50},
]

\addplot[
    color=green!60!black,
    dotted,
    mark=*,
    thick,
    mark size=1.0pt
] coordinates {
    (1,61.1) (2,22.9) (3,8.8) (4,3.5)
};


\addplot[
    color=orange!90!black,
    solid,
    mark=*,
    thick,
    mark size=1.0pt
] coordinates {
    (1,100.0) (2,99.5) (3,99.5) (4,100.0)
};

\end{axis}
\end{tikzpicture}
\caption{Bengali}
\label{fig:bn-gemma3-12b-it-combined}
\end{subfigure}

\vspace{0.5cm}

\begin{subfigure}[b]{0.24\textwidth}
\centering
\begin{tikzpicture}
\begin{axis}[
    width=\textwidth,
    height=0.9\textwidth,
    xlabel={Difficulty Level},
    ylabel={Score (\%)},
    xmin=0.5, xmax=4.5,
    ymin=-5, ymax=105,
    xtick={1,2,3,4},
    ytick={0,20,40,60,80,100},
    tick label style={font=\small},
    label style={font=\small},
    grid=major,
    grid style={line width=.1pt, draw=gray!30},
    major grid style={line width=.2pt, draw=gray!50},
]

\addplot[
    color=green!60!black,
    dotted,
    mark=*,
    thick,
    mark size=1.0pt
] coordinates {
    (1,81.9) (2,31.5) (3,15.2) (4,8.0)
};


\addplot[
    color=orange!90!black,
    solid,
    mark=*,
    thick,
    mark size=1.0pt
] coordinates {
    (1,98.9) (2,98.9) (3,98.7) (4,100.0)
};

\end{axis}
\end{tikzpicture}
\caption{French}
\label{fig:fr-gemma3-12b-it-combined}
\end{subfigure}
\hfill
\begin{subfigure}[b]{0.24\textwidth}
\centering
\begin{tikzpicture}
\begin{axis}[
    width=\textwidth,
    height=0.9\textwidth,
    xlabel={Difficulty Level},
    xmin=0.5, xmax=4.5,
    ymin=-5, ymax=105,
    xtick={1,2,3,4},
    ytick={0,20,40,60,80,100},
    tick label style={font=\small},
    label style={font=\small},
    grid=major,
    grid style={line width=.1pt, draw=gray!30},
    major grid style={line width=.2pt, draw=gray!50},
]

\addplot[
    color=green!60!black,
    dotted,
    mark=*,
    thick,
    mark size=1.0pt
] coordinates {
    (1,89.1) (2,29.6) (3,14.4) (4,8.0)
};


\addplot[
    color=orange!90!black,
    solid,
    mark=*,
    thick,
    mark size=1.0pt
] coordinates {
    (1,96.8) (2,99.5) (3,99.5) (4,99.7)
};

\end{axis}
\end{tikzpicture}
\caption{Russian}
\label{fig:ru-gemma3-12b-it-combined}
\end{subfigure}
\hfill
\begin{subfigure}[b]{0.24\textwidth}
\centering
\begin{tikzpicture}
\begin{axis}[
    width=\textwidth,
    height=0.9\textwidth,
    xlabel={Difficulty Level},
    xmin=0.5, xmax=4.5,
    ymin=-5, ymax=105,
    xtick={1,2,3,4},
    ytick={0,20,40,60,80,100},
    tick label style={font=\small},
    label style={font=\small},
    grid=major,
    grid style={line width=.1pt, draw=gray!30},
    major grid style={line width=.2pt, draw=gray!50},
]

\addplot[
    color=green!60!black,
    dotted,
    mark=*,
    thick,
    mark size=1.0pt
] coordinates {
    (1,77.9) (2,22.7) (3,9.9) (4,7.2)
};


\addplot[
    color=orange!90!black,
    solid,
    mark=*,
    thick,
    mark size=1.0pt
] coordinates {
    (1,97.9) (2,96.8) (3,97.1) (4,99.2)
};

\end{axis}
\end{tikzpicture}
\caption{Swahili}
\label{fig:sw-gemma3-12b-it-combined}
\end{subfigure}
\hfill
\begin{subfigure}[b]{0.24\textwidth}
\centering
\begin{tikzpicture}
\begin{axis}[
    width=\textwidth,
    height=0.9\textwidth,
    xlabel={Difficulty Level},
    xmin=0.5, xmax=4.5,
    ymin=-5, ymax=105,
    xtick={1,2,3,4},
    ytick={0,20,40,60,80,100},
    tick label style={font=\small},
    label style={font=\small},
    grid=major,
    grid style={line width=.1pt, draw=gray!30},
    major grid style={line width=.2pt, draw=gray!50},
]

\addplot[
    color=green!60!black,
    dotted,
    mark=*,
    thick,
    mark size=1.0pt
] coordinates {
    (1,74.9) (2,27.5) (3,10.4) (4,4.0)
};


\addplot[
    color=orange!90!black,
    solid,
    mark=*,
    thick,
    mark size=1.0pt
] coordinates {
    (1,100.0) (2,100.0) (3,99.2) (4,100.0)
};

\end{axis}
\end{tikzpicture}
\caption{Telugu}
\label{fig:te-gemma3-12b-it-combined}
\end{subfigure}

\captionof{figure}{
Difficulty-wise final-answer accuracy, thinking-language consistency (TC),
and answer-language consistency (AC) of Gemma-3-12B-IT across eight languages and four PolyMath difficulty levels: 1 = low, 2 = medium, 3 = high, and 4 = top.
The results show that answer-language
consistency is largely preserved across languages.
}
\label{fig:bf16_accuracy_tc_ac_difficulty_gemma3_12b_it}


\end{figure*}

\clearpage
\begin{figure*}[p]
\centering

\addtocounter{figure}{-1}

Phi-4-mini-instruct


\vspace{0.25cm}
\begin{tikzpicture}
\begin{axis}[
    hide axis,
    xmin=0, xmax=1, ymin=0, ymax=1,
    width=15cm,
    height=6cm,
    legend columns=3,
    legend style={
        draw=black!30,
        fill=white,
        font=\small,
        at={(0.5,0.5)},
        anchor=center,
        column sep=0.2cm,
    },
]
\addlegendimage{
    color=green!60!black,
    dotted,
    mark=*,
    thick,
    mark size=2pt
}
\addlegendentry{Accuracy}


\addlegendimage{
    color=orange!90!black,
    solid,
    mark=*,
    thick,
    mark size=2pt
}
\addlegendentry{Answer Consistency (AC)}
\end{axis}
\end{tikzpicture}

\vspace{0.4cm}

\begin{subfigure}[b]{0.24\textwidth}
\centering
\begin{tikzpicture}
\begin{axis}[
    width=\textwidth,
    height=0.9\textwidth,
    xlabel={Difficulty Level},
    ylabel={Score (\%)},
    xmin=0.5, xmax=4.5,
    ymin=-5, ymax=105,
    xtick={1,2,3,4},
    ytick={0,20,40,60,80,100},
    tick label style={font=\small},
    label style={font=\small},
    grid=major,
    grid style={line width=.1pt, draw=gray!30},
    major grid style={line width=.2pt, draw=gray!50},
]

\addplot[
    color=green!60!black,
    dotted,
    mark=*,
    thick,
    mark size=1.0pt
] coordinates {
    (1,88.0) (2,19.2) (3,4.8) (4,4.8)
};


\addplot[
    color=orange!90!black,
    solid,
    mark=*,
    thick,
    mark size=1.0pt
] coordinates {
    (1,100.0) (2,100.0) (3,100.0) (4,100.0)
};

\end{axis}
\end{tikzpicture}
\caption{English}
\label{fig:en-phi4mini-instruct-combined}
\end{subfigure}
\hfill
\begin{subfigure}[b]{0.24\textwidth}
\centering
\begin{tikzpicture}
\begin{axis}[
    width=\textwidth,
    height=0.9\textwidth,
    xlabel={Difficulty Level},
    xmin=0.5, xmax=4.5,
    ymin=-5, ymax=105,
    xtick={1,2,3,4},
    ytick={0,20,40,60,80,100},
    tick label style={font=\small},
    label style={font=\small},
    grid=major,
    grid style={line width=.1pt, draw=gray!30},
    major grid style={line width=.2pt, draw=gray!50},
]

\addplot[
    color=green!60!black,
    dotted,
    mark=*,
    thick,
    mark size=1.0pt
] coordinates {
    (1,56.0) (2,6.4) (3,2.4) (4,4.0)
};


\addplot[
    color=orange!90!black,
    solid,
    mark=*,
    thick,
    mark size=1.0pt
] coordinates {
    (1,100.0) (2,100.0) (3,100.0) (4,100.0)
};

\end{axis}
\end{tikzpicture}
\caption{Chinese}
\label{fig:zh-phi4mini-instruct-combined}
\end{subfigure}
\hfill
\begin{subfigure}[b]{0.24\textwidth}
\centering
\begin{tikzpicture}
\begin{axis}[
    width=\textwidth,
    height=0.9\textwidth,
    xlabel={Difficulty Level},
    xmin=0.5, xmax=4.5,
    ymin=-5, ymax=105,
    xtick={1,2,3,4},
    ytick={0,20,40,60,80,100},
    tick label style={font=\small},
    label style={font=\small},
    grid=major,
    grid style={line width=.1pt, draw=gray!30},
    major grid style={line width=.2pt, draw=gray!50},
]

\addplot[
    color=green!60!black,
    dotted,
    mark=*,
    thick,
    mark size=1.0pt
] coordinates {
    (1,51.2) (2,8.0) (3,2.4) (4,0.8)
};


\addplot[
    color=orange!90!black,
    solid,
    mark=*,
    thick,
    mark size=1.0pt
] coordinates {
    (1,99.2) (2,99.2) (3,99.2) (4,100.0)
};

\end{axis}
\end{tikzpicture}
\caption{Arabic}
\label{fig:ar-phi4mini-instruct-combined}
\end{subfigure}
\hfill
\begin{subfigure}[b]{0.24\textwidth}
\centering
\begin{tikzpicture}
\begin{axis}[
    width=\textwidth,
    height=0.9\textwidth,
    xlabel={Difficulty Level},
    xmin=0.5, xmax=4.5,
    ymin=-5, ymax=105,
    xtick={1,2,3,4},
    ytick={0,20,40,60,80,100},
    tick label style={font=\small},
    label style={font=\small},
    grid=major,
    grid style={line width=.1pt, draw=gray!30},
    major grid style={line width=.2pt, draw=gray!50},
]

\addplot[
    color=green!60!black,
    dotted,
    mark=*,
    thick,
    mark size=1.0pt
] coordinates {
    (1,22.4) (2,4.8) (3,0.8) (4,1.6)
};


\addplot[
    color=orange!90!black,
    solid,
    mark=*,
    thick,
    mark size=1.0pt
] coordinates {
    (1,94.4) (2,88.0) (3,94.4) (4,87.2)
};

\end{axis}
\end{tikzpicture}
\caption{Bengali}
\label{fig:bn-phi4mini-instruct-combined}
\end{subfigure}

\vspace{0.5cm}

\begin{subfigure}[b]{0.24\textwidth}
\centering
\begin{tikzpicture}
\begin{axis}[
    width=\textwidth,
    height=0.9\textwidth,
    xlabel={Difficulty Level},
    ylabel={Score (\%)},
    xmin=0.5, xmax=4.5,
    ymin=-5, ymax=105,
    xtick={1,2,3,4},
    ytick={0,20,40,60,80,100},
    tick label style={font=\small},
    label style={font=\small},
    grid=major,
    grid style={line width=.1pt, draw=gray!30},
    major grid style={line width=.2pt, draw=gray!50},
]

\addplot[
    color=green!60!black,
    dotted,
    mark=*,
    thick,
    mark size=1.0pt
] coordinates {
    (1,70.4) (2,13.6) (3,2.4) (4,3.2)
};


\addplot[
    color=orange!90!black,
    solid,
    mark=*,
    thick,
    mark size=1.0pt
] coordinates {
    (1,99.2) (2,100.0) (3,100.0) (4,100.0)
};

\end{axis}
\end{tikzpicture}
\caption{French}
\label{fig:fr-phi4mini-instruct-combined}
\end{subfigure}
\hfill
\begin{subfigure}[b]{0.24\textwidth}
\centering
\begin{tikzpicture}
\begin{axis}[
    width=\textwidth,
    height=0.9\textwidth,
    xlabel={Difficulty Level},
    xmin=0.5, xmax=4.5,
    ymin=-5, ymax=105,
    xtick={1,2,3,4},
    ytick={0,20,40,60,80,100},
    tick label style={font=\small},
    label style={font=\small},
    grid=major,
    grid style={line width=.1pt, draw=gray!30},
    major grid style={line width=.2pt, draw=gray!50},
]

\addplot[
    color=green!60!black,
    dotted,
    mark=*,
    thick,
    mark size=1.0pt
] coordinates {
    (1,68.0) (2,8.0) (3,3.2) (4,2.4)
};


\addplot[
    color=orange!90!black,
    solid,
    mark=*,
    thick,
    mark size=1.0pt
] coordinates {
    (1,98.4) (2,100.0) (3,99.2) (4,100.0)
};

\end{axis}
\end{tikzpicture}
\caption{Russian}
\label{fig:ru-phi4mini-instruct-combined}
\end{subfigure}
\hfill
\begin{subfigure}[b]{0.24\textwidth}
\centering
\begin{tikzpicture}
\begin{axis}[
    width=\textwidth,
    height=0.9\textwidth,
    xlabel={Difficulty Level},
    xmin=0.5, xmax=4.5,
    ymin=-5, ymax=105,
    xtick={1,2,3,4},
    ytick={0,20,40,60,80,100},
    tick label style={font=\small},
    label style={font=\small},
    grid=major,
    grid style={line width=.1pt, draw=gray!30},
    major grid style={line width=.2pt, draw=gray!50},
]

\addplot[
    color=green!60!black,
    dotted,
    mark=*,
    thick,
    mark size=1.0pt
] coordinates {
    (1,18.4) (2,2.4) (3,0.0) (4,3.2)
};


\addplot[
    color=orange!90!black,
    solid,
    mark=*,
    thick,
    mark size=1.0pt
] coordinates {
    (1,96.8) (2,92.0) (3,88.0) (4,93.6)
};

\end{axis}
\end{tikzpicture}
\caption{Swahili}
\label{fig:sw-phi4mini-instruct-combined}
\end{subfigure}
\hfill
\begin{subfigure}[b]{0.24\textwidth}
\centering
\begin{tikzpicture}
\begin{axis}[
    width=\textwidth,
    height=0.9\textwidth,
    xlabel={Difficulty Level},
    xmin=0.5, xmax=4.5,
    ymin=-5, ymax=105,
    xtick={1,2,3,4},
    ytick={0,20,40,60,80,100},
    tick label style={font=\small},
    label style={font=\small},
    grid=major,
    grid style={line width=.1pt, draw=gray!30},
    major grid style={line width=.2pt, draw=gray!50},
]

\addplot[
    color=green!60!black,
    dotted,
    mark=*,
    thick,
    mark size=1.0pt
] coordinates {
    (1,17.6) (2,4.8) (3,3.2) (4,0.8)
};


\addplot[
    color=orange!90!black,
    solid,
    mark=*,
    thick,
    mark size=1.0pt
] coordinates {
    (1,100.0) (2,97.6) (3,97.6) (4,98.4)
};

\end{axis}
\end{tikzpicture}
\caption{Telugu}
\label{fig:te-phi4mini-instruct-combined}
\end{subfigure}

\captionof{figure}{
Difficulty-wise final-answer accuracy, thinking-language consistency (TC),
and answer-language consistency (AC) of Phi-4-mini-instruct across eight languages and four PolyMath difficulty levels: 1 = low, 2 = medium, 3 = high, and 4 = top.
The results show that answer-language consistency remains stable across most languages but decreases for a few as task difficulty increases.
}
\label{fig:bf16_accuracy_tc_ac_difficulty_phi4mini_instruct}


\vspace{0.4cm}

Phi-4-mini-reasoning


\vspace{0.25cm}

\begin{tikzpicture}
\begin{axis}[
    hide axis,
    xmin=0, xmax=1, ymin=0, ymax=1,
    width=15cm,
    height=6cm,
    legend columns=3,
    legend style={
        draw=black!30,
        fill=white,
        font=\small,
        at={(0.5,0.5)},
        anchor=center,
        column sep=0.2cm,
    },
]
\addlegendimage{
    color=green!60!black,
    dotted,
    mark=*,
    thick,
    mark size=2pt
}
\addlegendentry{Accuracy}

\addlegendimage{
    color=orange!90!black,
    solid,
    mark=*,
    thick,
    mark size=2pt
}
\addlegendentry{Answer Consistency (AC)}

\addlegendimage{
    color=red!85!black,
    solid,
    mark=*,
    thick,
    mark size=2pt
}
\addlegendentry{Thinking Consistency (TC)}

\end{axis}
\end{tikzpicture}

\vspace{0.4cm}

\begin{subfigure}[b]{0.24\textwidth}
\centering
\begin{tikzpicture}
\begin{axis}[
    width=\textwidth,
    height=0.9\textwidth,
    xlabel={Difficulty Level},
    ylabel={Score (\%)},
    xmin=0.5, xmax=4.5,
    ymin=-5, ymax=105,
    xtick={1,2,3,4},
    ytick={0,20,40,60,80,100},
    tick label style={font=\small},
    label style={font=\small},
    grid=major,
    grid style={line width=.1pt, draw=gray!30},
    major grid style={line width=.2pt, draw=gray!50},
]

\addplot[
    color=green!60!black,
    dotted,
    mark=*,
    thick,
    mark size=1.0pt
] coordinates {
    (1,94.0) (2,54.0) (3,27.6) (4,14.0)
};

\addplot[
    color=red!85!black,
    solid,
    mark=*,
    thick,
    mark size=1.0pt
] coordinates {
    (1,100.0) (2,99.2) (3,99.6) (4,100.0)
};

\addplot[
    color=orange!90!black,
    solid,
    mark=*,
    thick,
    mark size=1.0pt
] coordinates {
    (1,100.0) (2,100.0) (3,99.6) (4,100.0)
};

\end{axis}
\end{tikzpicture}
\caption{English}
\label{fig:en-phi4mini-reasoning-combined}
\end{subfigure}
\hfill
\begin{subfigure}[b]{0.24\textwidth}
\centering
\begin{tikzpicture}
\begin{axis}[
    width=\textwidth,
    height=0.9\textwidth,
    xlabel={Difficulty Level},
    xmin=0.5, xmax=4.5,
    ymin=-5, ymax=105,
    xtick={1,2,3,4},
    ytick={0,20,40,60,80,100},
    tick label style={font=\small},
    label style={font=\small},
    grid=major,
    grid style={line width=.1pt, draw=gray!30},
    major grid style={line width=.2pt, draw=gray!50},
]

\addplot[
    color=green!60!black,
    dotted,
    mark=*,
    thick,
    mark size=1.0pt
] coordinates {
    (1,79.6) (2,39.6) (3,18.0) (4,5.6)
};

\addplot[
    color=red!85!black,
    solid,
    mark=*,
    thick,
    mark size=1.0pt
] coordinates {
    (1,100) (2,99.2) (3,99.6) (4,100)
};

\addplot[
    color=orange!90!black,
    solid,
    mark=*,
    thick,
    mark size=1.0pt
] coordinates {
    (1,100.0) (2,100.0) (3,99.6) (4,100.0)
};

\end{axis}
\end{tikzpicture}
\caption{Chinese}
\label{fig:zh-phi4mini-reasoning-combined}
\end{subfigure}
\hfill
\begin{subfigure}[b]{0.24\textwidth}
\centering
\begin{tikzpicture}
\begin{axis}[
    width=\textwidth,
    height=0.9\textwidth,
    xlabel={Difficulty Level},
    xmin=0.5, xmax=4.5,
    ymin=-5, ymax=105,
    xtick={1,2,3,4},
    ytick={0,20,40,60,80,100},
    tick label style={font=\small},
    label style={font=\small},
    grid=major,
    grid style={line width=.1pt, draw=gray!30},
    major grid style={line width=.2pt, draw=gray!50},
]

\addplot[
    color=green!60!black,
    dotted,
    mark=*,
    thick,
    mark size=1.0pt
] coordinates {
    (1,72.4) (2,43.2) (3,23.2) (4,9.6)
};

\addplot[
    color=red!85!black,
    solid,
    mark=*,
    thick,
    mark size=1.0pt
] coordinates {
    (1,2.8) (2,9.2) (3,8.8) (4,12.8)
};

\addplot[
    color=orange!90!black,
    solid,
    mark=*,
    thick,
    mark size=1.0pt
] coordinates {
    (1,64.0) (2,8.0) (3,4.0) (4,3.6)
};

\end{axis}
\end{tikzpicture}
\caption{Arabic}
\label{fig:ar-phi4mini-reasoning-combined}
\end{subfigure}
\hfill
\begin{subfigure}[b]{0.24\textwidth}
\centering
\begin{tikzpicture}
\begin{axis}[
    width=\textwidth,
    height=0.9\textwidth,
    xlabel={Difficulty Level},
    xmin=0.5, xmax=4.5,
    ymin=-5, ymax=105,
    xtick={1,2,3,4},
    ytick={0,20,40,60,80,100},
    tick label style={font=\small},
    label style={font=\small},
    grid=major,
    grid style={line width=.1pt, draw=gray!30},
    major grid style={line width=.2pt, draw=gray!50},
]

\addplot[
    color=green!60!black,
    dotted,
    mark=*,
    thick,
    mark size=1.0pt
] coordinates {
    (1,30.8) (2,28.0) (3,16.8) (4,8.8)
};

\addplot[
    color=red!85!black,
    solid,
    mark=*,
    thick,
    mark size=1.0pt
] coordinates {
    (1,20.4) (2,10.0) (3,14.4) (4,21.2)
};

\addplot[
    color=orange!90!black,
    solid,
    mark=*,
    thick,
    mark size=1.0pt
] coordinates {
    (1,39.2) (2,14.0) (3,2.8) (4,5.2)
};

\end{axis}
\end{tikzpicture}
\caption{Bengali}
\label{fig:bn-phi4mini-reasoning-combined}
\end{subfigure}

\vspace{0.5cm}

\begin{subfigure}[b]{0.24\textwidth}
\centering
\begin{tikzpicture}
\begin{axis}[
    width=\textwidth,
    height=0.9\textwidth,
    xlabel={Difficulty Level},
    ylabel={Score (\%)},
    xmin=0.5, xmax=4.5,
    ymin=-5, ymax=105,
    xtick={1,2,3,4},
    ytick={0,20,40,60,80,100},
    tick label style={font=\small},
    label style={font=\small},
    grid=major,
    grid style={line width=.1pt, draw=gray!30},
    major grid style={line width=.2pt, draw=gray!50},
]

\addplot[
    color=green!60!black,
    dotted,
    mark=*,
    thick,
    mark size=1.0pt
] coordinates {
    (1,83.2) (2,50.8) (3,29.2) (4,15.6)
};

\addplot[
    color=red!85!black,
    solid,
    mark=*,
    thick,
    mark size=1.0pt
] coordinates {
    (1,4.4) (2,6.4) (3,12.4) (4,14.0)
};

\addplot[
    color=orange!90!black,
    solid,
    mark=*,
    thick,
    mark size=1.0pt
] coordinates {
    (1,96.4) (2,40.0) (3,32.0) (4,40.0)
};

\end{axis}
\end{tikzpicture}
\caption{French}
\label{fig:fr-phi4mini-reasoning-combined}
\end{subfigure}
\hfill
\begin{subfigure}[b]{0.24\textwidth}
\centering
\begin{tikzpicture}
\begin{axis}[
    width=\textwidth,
    height=0.9\textwidth,
    xlabel={Difficulty Level},
    xmin=0.5, xmax=4.5,
    ymin=-5, ymax=105,
    xtick={1,2,3,4},
    ytick={0,20,40,60,80,100},
    tick label style={font=\small},
    label style={font=\small},
    grid=major,
    grid style={line width=.1pt, draw=gray!30},
    major grid style={line width=.2pt, draw=gray!50},
]

\addplot[
    color=green!60!black,
    dotted,
    mark=*,
    thick,
    mark size=1.0pt
] coordinates {
    (1,80.8) (2,37.6) (3,15.6) (4,3.2)
};

\addplot[
    color=red!85!black,
    solid,
    mark=*,
    thick,
    mark size=1.0pt
] coordinates {
    (1,51.6) (2,64.0) (3,62.8) (4,66.0)
};

\addplot[
    color=orange!90!black,
    solid,
    mark=*,
    thick,
    mark size=1.0pt
] coordinates {
    (1,89.2) (2,54.4) (3,63.2) (4,71.2)
};

\end{axis}
\end{tikzpicture}
\caption{Russian}
\label{fig:ru-phi4mini-reasoning-combined}
\end{subfigure}
\hfill
\begin{subfigure}[b]{0.24\textwidth}
\centering
\begin{tikzpicture}
\begin{axis}[
    width=\textwidth,
    height=0.9\textwidth,
    xlabel={Difficulty Level},
    xmin=0.5, xmax=4.5,
    ymin=-5, ymax=105,
    xtick={1,2,3,4},
    ytick={0,20,40,60,80,100},
    tick label style={font=\small},
    label style={font=\small},
    grid=major,
    grid style={line width=.1pt, draw=gray!30},
    major grid style={line width=.2pt, draw=gray!50},
]

\addplot[
    color=green!60!black,
    dotted,
    mark=*,
    thick,
    mark size=1.0pt
] coordinates {
    (1,33.2) (2,28.4) (3,17.6) (4,6.8)
};

\addplot[
    color=red!85!black,
    solid,
    mark=*,
    thick,
    mark size=1.0pt
] coordinates {
    (1,5.6) (2,4.8) (3,10.4) (4,9.2)
};

\addplot[
    color=orange!90!black,
    solid,
    mark=*,
    thick,
    mark size=1.0pt
] coordinates {
    (1,82.0) (2,31.6) (3,26.0) (4,24.4)
};

\end{axis}
\end{tikzpicture}
\caption{Swahili}
\label{fig:sw-phi4mini-reasoning-combined}
\end{subfigure}
\hfill
\begin{subfigure}[b]{0.24\textwidth}
\centering
\begin{tikzpicture}
\begin{axis}[
    width=\textwidth,
    height=0.9\textwidth,
    xlabel={Difficulty Level},
    xmin=0.5, xmax=4.5,
    ymin=-5, ymax=105,
    xtick={1,2,3,4},
    ytick={0,20,40,60,80,100},
    tick label style={font=\small},
    label style={font=\small},
    grid=major,
    grid style={line width=.1pt, draw=gray!30},
    major grid style={line width=.2pt, draw=gray!50},
]

\addplot[
    color=green!60!black,
    dotted,
    mark=*,
    thick,
    mark size=1.0pt
] coordinates {
    (1,21.6) (2,30.4) (3,15.2) (4,6.4)
};

\addplot[
    color=red!85!black,
    solid,
    mark=*,
    thick,
    mark size=1.0pt
] coordinates {
    (1,18.0) (2,6.4) (3,14.4) (4,17.2)
};

\addplot[
    color=orange!90!black,
    solid,
    mark=*,
    thick,
    mark size=1.0pt
] coordinates {
    (1,19.6) (2,3.6) (3,1.6) (4,0.8)
};

\end{axis}
\end{tikzpicture}
\caption{Telugu}
\label{fig:te-phi4mini-reasoning-combined}
\end{subfigure}

\captionof{figure}{
Difficulty-wise final-answer accuracy, thinking-language consistency (TC),
and answer-language consistency (AC) of Phi-4-mini-reasoning across eight languages and four PolyMath difficulty levels: 1 = low, 2 = medium, 3 = high, and 4 = top.
The results show that thinking- and answer-language consistency are preserved for English and Chinese, while answer-language consistency breaks down for several other languages.
}
\label{fig:bf16_accuracy_tc_ac_difficulty_phi4mini_reasoning}


\end{figure*}

\clearpage

\begin{table*}[!t]
\centering
\renewcommand{\arraystretch}{1.20}
\resizebox{1.0\linewidth}{!}{
\begin{tabular}{llcccll}
\hline\hline
\textbf{Model} &
\textbf{Method} &
\textbf{W-A-KV Bits} &
\textbf{Avg. Acc.} &
\textbf{Avg. TC} &
\textbf{Accuracy-wise Winner Lang.} &
\textbf{Accuracy-wise Loss Lang.} \\
\hline

\multirow{4}{*}{\textbf{\shortstack[l]{DeepSeek-R1\\Distill-Qwen-7B}}}
& --- & \textbf{---} & \textbf{36.5\small{$\pm$1.5}} & \textbf{52.3\small{$\pm$3.0}} & -- & -- \\
\cline{2-7}
& GPTQ & 4-16-16 & \cellcolor{red!15}33.1\small{$\pm$0.9} & 46.9\small{$\pm$2.1} & -- &
\cellcolor{red!15}{en, zh, ar, bn, sw, te} \\
& AWQ & 4-16-16 & 33.9\small{$\pm$1.4} & \cellcolor{green!15}48.5\small{$\pm$3.2} &
bn, sw & fr, ru \\
& AutoRound & 4-16-16 & \cellcolor{green!15}35.7\small{$\pm$1.4} &
\cellcolor{red!15}45.6\small{$\pm$1.2} &
\cellcolor{green!15}{en, zh, ar, fr, ru, te} & -- \\
\hline

\multirow{4}{*}{\textbf{OLMo-3-7B-Think}}
& --- & \textbf{---} & \textbf{42.7\small{$\pm$1.0}} & \textbf{27.0\small{$\pm$0.4}} & -- & -- \\
\cline{2-7}
& GPTQ & 4-16-16 & 40.4\small{$\pm$0.9} & \cellcolor{green!15}31.3\small{$\pm$0.6} &
en & \cellcolor{red!15}{zh, ar, sw, te} \\
& AWQ & 4-16-16 & \cellcolor{red!15}40.3\small{$\pm$1.4} & 29.1\small{$\pm$0.6} &
en & bn, fr, ru \\
& AutoRound & 4-16-16 & \cellcolor{green!15}41.6\small{$\pm$1.2} &
\cellcolor{red!15}26.8\small{$\pm$0.4} &
\cellcolor{green!15}{zh, ar, bn, fr, ru, sw, te} & en \\
\hline\hline
\end{tabular}
}
\caption{
Summary of accuracy and thinking-language consistency (TC) across eight
languages for DeepSeek-R1-Distill-Qwen-7B and OLMo-3-7B-Think.
Avg. Acc. and Avg. TC report averages over the eight shown languages.
Green/red cells mark the best/worst values among GPTQ, AWQ, and AutoRound
within each model.
Accuracy-wise winner/loss languages indicate where each method obtains the
highest/lowest accuracy within the same model.
Full per-language Acc./TC results are reported in
Table~\ref{tab:quant-acc-tc-scores}.
}
\vspace{-2ex}
\label{tab:quant-acc-tc-scores-compact}
\end{table*}

\begin{table*}[!t]
\begin{center}
\renewcommand{\arraystretch}{1.8}
\resizebox{1.0\linewidth}{!}{
\begin{tabular}{c|c|rr|rr|rr|rr|rr|rr|rr|rr|rr}
\hline\hline
\multirow{2}{*}{\textbf{Model}} &
\multirow{2}{*}{\textbf{Method}} &
\multicolumn{2}{c|}{\textbf{en}} &
\multicolumn{2}{c|}{\textbf{zh}} &
\multicolumn{2}{c|}{\textbf{ar}} &
\multicolumn{2}{c|}{\textbf{bn}} &
\multicolumn{2}{c|}{\textbf{fr}} &
\multicolumn{2}{c|}{\textbf{ru}} &
\multicolumn{2}{c|}{\textbf{sw}} &
\multicolumn{2}{c|}{\textbf{te}} &
\multicolumn{2}{c}{\textbf{Avg}} \\
\cline{3-20}
& &
\textbf{Acc.} & \textbf{TC} &
\textbf{Acc.} & \textbf{TC} &
\textbf{Acc.} & \textbf{TC} &
\textbf{Acc.} & \textbf{TC} &
\textbf{Acc.} & \textbf{TC} &
\textbf{Acc.} & \textbf{TC} &
\textbf{Acc.} & \textbf{TC} &
\textbf{Acc.} & \textbf{TC} &
\textbf{Acc.} & \textbf{TC} \\
\hline
    \multirow{3}{*}[-3ex]{\rotatebox[origin=c]{90}{\textbf{\makecell{DS-R1\\Qwen-7B}}}}
    & ---
    & \textbf{49.2\small{$\pm$1.7}} & \textbf{99.8\small{$\pm$0.2}}
    & \textbf{46.0\small{$\pm$0.5}} & \textbf{92.7\small{$\pm$0.9}}
    & \textbf{39.2\small{$\pm$2.8}} & \textbf{20.2\small{$\pm$1.1}}
    & \textbf{36.3\small{$\pm$1.0}} & \textbf{26.2\small{$\pm$0.2}}
    & \textbf{42.0\small{$\pm$0.6}} & \textbf{37.8\small{$\pm$10}}
    & \textbf{37.6\small{$\pm$1.7}} & \textbf{88.9\small{$\pm$1.4}}
    & \textbf{16.1\small{$\pm$1.4}} & \textbf{26.0\small{$\pm$2.9}}
    & \textbf{25.3\small{$\pm$2.6}} & \textbf{31.1\small{$\pm$7.1}}
    & \textbf{36.5\small{$\pm$1.5}} & \textbf{52.3\small{$\pm$3.0}} \\
    \cline{2-20}
    & GPTQ
    & \cellcolor{red!15}44.8\small{$\pm$2.0} & 99.7\small{$\pm$0.2}
    & \cellcolor{red!15}40.3\small{$\pm$0.2} & 94.8\small{$\pm$1.2}
    & \cellcolor{red!15}33.6\small{$\pm$1.6} & 14.5\small{$\pm$1.1}
    & \cellcolor{red!15}32.4\small{$\pm$1.2} & 19.6\small{$\pm$2.7}
    & 39.8\small{$\pm$0.3} & 31.0\small{$\pm$3.8}
    & 37.0\small{$\pm$0.7} & 61.4\small{$\pm$3.0}
    & \cellcolor{red!15}14.1\small{$\pm$0.5} & 27.3\small{$\pm$2.8}
    & \cellcolor{red!15}22.6\small{$\pm$0.7} & 31.2\small{$\pm$1.7}
    & \cellcolor{red!15}33.1\small{$\pm$0.9} & 46.9\small{$\pm$2.1} \\
    & AWQ
    & 45.6\small{$\pm$0.2} & 99.7\small{$\pm$0.3}
    & 42.0\small{$\pm$1.7} & 92.7\small{$\pm$1.0}
    & 36.3\small{$\pm$2.5} & 15.5\small{$\pm$1.6}
    & \cellcolor{green!15}33.7\small{$\pm$0.5} & 21.8\small{$\pm$2.9}
    & \cellcolor{red!15}39.5\small{$\pm$1.9} & 30.2\small{$\pm$3.8}
    & \cellcolor{red!15}34.8\small{$\pm$1.6} & 86.6\small{$\pm$1.7}
    & \cellcolor{green!15}16.0\small{$\pm$0.6} & 20.4\small{$\pm$8.3}
    & 23.6\small{$\pm$2.5} & 25.2\small{$\pm$6.0}
    & 33.9\small{$\pm$1.4} & 48.5\small{$\pm$3.2} \\
    & AutoRound
    & \cellcolor{green!15}48.0\small{$\pm$2.8} & 100\small{$\pm$0.0}
    & \cellcolor{green!15}43.3\small{$\pm$2.1} & 91.9\small{$\pm$0.8}
    & \cellcolor{green!15}38.8\small{$\pm$2.0} & 14.5\small{$\pm$0.6}
    & 32.7\small{$\pm$0.7} & 21.3\small{$\pm$1.5}
    & \cellcolor{green!15}41.8\small{$\pm$0.6} & 26.8\small{$\pm$0.9}
    & \cellcolor{green!15}40.2\small{$\pm$1.6} & 63.5\small{$\pm$0.9}
    & 15.0\small{$\pm$1.2} & 25.8\small{$\pm$3.1}
    & \cellcolor{green!15}25.5\small{$\pm$0.5} & 24.9\small{$\pm$1.8}
    & \cellcolor{green!15}35.7\small{$\pm$1.4} & 45.6\small{$\pm$1.2} \\
    \hline\hline

    \multirow{3}{*}[-3ex]{\rotatebox[origin=c]{90}{\textbf{\makecell{Olmo-3-7B\\-Think}}}}
    & ---
    & \textbf{57.6}\small{$\pm$1.2} & \textbf{99.7}\small{$\pm$0.1} 
    & \textbf{54.0}\small{$\pm$1.2} & \textbf{11.3}\small{$\pm$0.5} 
    & \textbf{44.7}\small{$\pm$1.3} & \textbf{0.5}\small{$\pm$0.3}  
    & \textbf{32.9}\small{$\pm$1.3} & \textbf{3.1}\small{$\pm$1.0}  
    & \textbf{53.9}\small{$\pm$0.5} & \textbf{0.4}\small{$\pm$0.1}  
    & \textbf{38.1}\small{$\pm$1.4} & \textbf{99.4}\small{$\pm$0.6} 
    & \textbf{22.9}\small{$\pm$0.5} & \textbf{0.7}\small{$\pm$0.3}  
    & \textbf{37.6}\small{$\pm$0.9} & \textbf{1.1}\small{$\pm$0.6}  
    & \textbf{42.7}\small{$\pm$1.0} & \textbf{27.0}\small{$\pm$0.4} \\ 
    \cline{2-20}
    & GPTQ
    & \cellcolor{green!15}57.4\small{$\pm$0.4} & 99.9\small{$\pm$0.1}
    & \cellcolor{red!15}49.3\small{$\pm$1.9} & 44.2\small{$\pm$1.6}
    & \cellcolor{red!15}43.0\small{$\pm$1.1} & 1.3\small{$\pm$0.4}
    & 30.1\small{$\pm$1.1} & 3.5\small{$\pm$0.4}
    & 52.9\small{$\pm$0.8} & 0.6\small{$\pm$0.4}
    & 35.6\small{$\pm$0.9} & 98.2\small{$\pm$0.7}
    & \cellcolor{red!15}20.4\small{$\pm$0.4} & 1.3\small{$\pm$0.6}
    & \cellcolor{red!15}34.4\small{$\pm$0.8} & 1.4\small{$\pm$0.3}
    & 40.4\small{$\pm$0.9} & 31.3\small{$\pm$0.6} \\
    & AWQ
    & \cellcolor{green!15}57.4\small{$\pm$0.5} & 99.7\small{$\pm$0.1}
    & 51.0\small{$\pm$1.9} & 24.2\small{$\pm$1.9}
    & 43.8\small{$\pm$0.5} & 1.2\small{$\pm$0.3}
    & \cellcolor{red!15}29.2\small{$\pm$2.3} & 4.3\small{$\pm$0.7}
    & \cellcolor{red!15}51.2\small{$\pm$0.8} & 1.6\small{$\pm$1.0}
    & \cellcolor{red!15}34.1\small{$\pm$1.6} & 98.3\small{$\pm$0.2}
    & 20.7\small{$\pm$1.0} & 1.0\small{$\pm$0.2}
    & 35.2\small{$\pm$2.6} & 2.2\small{$\pm$0.8}
    & \cellcolor{red!15}40.3\small{$\pm$1.4} & 29.1\small{$\pm$0.6} \\
    & AutoRound
    & \cellcolor{red!15}56.2\small{$\pm$0.9} & 99.8\small{$\pm$0.2}
    & \cellcolor{green!15}51.5\small{$\pm$1.0} & 11.6\small{$\pm$1.3}
    & \cellcolor{green!15}46.0\small{$\pm$2.0} & 0.6\small{$\pm$0.2}
    & \cellcolor{green!15}30.5\small{$\pm$1.7} & 2.7\small{$\pm$0.4}
    & \cellcolor{green!15}54.2\small{$\pm$1.2} & 1.1\small{$\pm$0.4}
    & \cellcolor{green!15}37.5\small{$\pm$0.3} & 96.9\small{$\pm$0.5}
    & \cellcolor{green!15}21.1\small{$\pm$0.8} & 0.7\small{$\pm$0.3}
    & \cellcolor{green!15}35.6\small{$\pm$0.8} & 1.3\small{$\pm$0.3}
    & \cellcolor{green!15}41.6\small{$\pm$1.2} & 26.8\small{$\pm$0.4} \\
    \hline\hline
\end{tabular}
}
\end{center}
\caption{
Accuracy and thinking-language consistency (TC) results across eight languages and quantization methods. 
\textbf{Acc.} denotes benchmark accuracy and \textbf{TC} denotes thinking-language consistency scores. 
The \textcolor{green!50}{green}/\textcolor{red!50}{red} cells mark the best/worst accuracy values respectively within each model-language-quantization group and total average column.
}
\vspace{-2ex}
\label{tab:quant-acc-tc-scores}
\end{table*}

\begin{table*}[!t]
\begin{center}
\renewcommand{\arraystretch}{1.8}
\resizebox{1.0\linewidth}{!}{
\begin{tabular}{c|c|c|rr|rr|rr|rr|rr|rr|rr|rr|rr}
\hline\hline
\multirow{2}{*}{\textbf{Model}} &
\multirow{2}{*}{\textbf{Method}} &
\multirow{2}{*}{\textbf{\makecell{W-A-KV\\\# Bits}}} &
\multicolumn{2}{c|}{\textbf{en}} &
\multicolumn{2}{c|}{\textbf{zh}} &
\multicolumn{2}{c|}{\textbf{ar}} &
\multicolumn{2}{c|}{\textbf{bn}} &
\multicolumn{2}{c|}{\textbf{fr}} &
\multicolumn{2}{c|}{\textbf{ru}} &
\multicolumn{2}{c|}{\textbf{sw}} &
\multicolumn{2}{c|}{\textbf{te}} &
\multicolumn{2}{c}{\textbf{\makecell{Avg.\\$\Delta$}}} \\
\cline{4-21}
& & &
$\Delta$\textbf{Acc.} & $\Delta$\textbf{TC} &
$\Delta$\textbf{Acc.} & $\Delta$\textbf{TC} &
$\Delta$\textbf{Acc.} & $\Delta$\textbf{TC} &
$\Delta$\textbf{Acc.} & $\Delta$\textbf{TC} &
$\Delta$\textbf{Acc.} & $\Delta$\textbf{TC} &
$\Delta$\textbf{Acc.} & $\Delta$\textbf{TC} &
$\Delta$\textbf{Acc.} & $\Delta$\textbf{TC} &
$\Delta$\textbf{Acc.} & $\Delta$\textbf{TC} &
\textbf{Acc.} & \textbf{TC} \\
\hline
    \multirow{3}{*}[-2ex]{\rotatebox[origin=c]{90}{\textbf{\makecell{DS-R1\\Qwen-7B}}}}
    & GPTQ
    & 4-16-16
    & -4.4 & \cellcolor{yellow!15}-0.1
    & -5.7 & \cellcolor{yellow!15}+2.1
    & -5.6 & \cellcolor{yellow!15}-5.7
    & -3.9 & -6.6
    & -2.2 & \cellcolor{yellow!15}-6.8
    & -0.6 & -27.5
    & -2.0 & \cellcolor{yellow!15}+1.3
    & -2.7 & \cellcolor{yellow!15}+0.1
    & -3.39 & -5.40 \\
    & AWQ
    & 4-16-16
    & -3.6 & \cellcolor{yellow!15}-0.1
    & -4.0 & 0.0
    & -2.9 & \cellcolor{yellow!15}-4.7
    & -2.6 & \cellcolor{yellow!15}-4.4
    & -2.5 & \cellcolor{yellow!15}-7.6
    & -2.8 & \cellcolor{yellow!15}-2.3
    & -0.1 & -5.6
    & -1.7 & -5.9
    & -2.53 & \cellcolor{green!15}-3.83 \\
    & AutoRound
    & 4-16-16
    & -1.2 & \cellcolor{yellow!15}+0.2
    & -2.7 & -0.8
    & -0.4 & \cellcolor{yellow!15}-5.7
    & -3.6 & \cellcolor{yellow!15}-4.9
    & -0.2 & -11.0
    & +2.6 & -25.4
    & -1.1 & -0.2
    & +0.2 & -6.2
    & -0.80 & \cellcolor{red!15}-6.75 \\
    \hline\hline

    \multirow{3}{*}[-2ex]{\rotatebox[origin=c]{90}{\textbf{\makecell{Olmo-3-7B\\-Think}}}}
    & GPTQ
    & 4-16-16
    & -0.2 & \cellcolor{yellow!15}+0.2
    & -4.7 & \cellcolor{yellow!15}+32.9
    & -1.7 & \cellcolor{yellow!15}+0.8
    & -2.8 & \cellcolor{yellow!15}+0.4
    & -1.0 & +0.2
    & -2.5 & \cellcolor{yellow!15}-1.2
    & -2.5 & \cellcolor{yellow!15}+0.6
    & -3.2 & \cellcolor{yellow!15}+0.3
    & -2.33 & \cellcolor{green!15}+4.28 \\
    & AWQ
    & 4-16-16
    & -0.2 & \cellcolor{yellow!15}0.0
    & -3.0 & +12.9
    & -0.9 & \cellcolor{yellow!15}+0.7
    & -3.7 & \cellcolor{yellow!15}+1.2
    & -2.7 & \cellcolor{yellow!15}+1.2
    & -4.0 & \cellcolor{yellow!15}-1.1
    & -2.2 & \cellcolor{yellow!15}+0.3
    & -2.4 & \cellcolor{yellow!15}+1.1
    & -2.39 & +2.04 \\
    & AutoRound
    & 4-16-16
    & -1.4 & \cellcolor{yellow!15}+0.1
    & -2.5 & +0.3
    & +1.3 & \cellcolor{yellow!15}+0.1
    & -2.4 & -0.4
    & +0.3 & \cellcolor{yellow!15}+0.7
    & -0.6 & -2.5
    & -1.8 & \cellcolor{yellow!15}0.0
    & -2.0 & \cellcolor{yellow!15}+0.2
    & -1.14 & \cellcolor{red!15}-0.19 \\
    \hline\hline
\end{tabular}
}
\end{center}
\caption{
Accuracy and Thinking-language Consistency (TC) $\Delta$ scores, relative to Baseline across eight languages. 
The \textcolor{green!50}{green}/\textcolor{red!50}{red} cells mark the best/worst integer methods by magnitude-based $\Delta$TC scores.
\textcolor{darkyellow}{Yellow} cells indicate tolerance-based TC voting co-winners with $\epsilon=1.0$, i.e., methods within one TC point of the best $\Delta$TC for each model-language-quantization group are considered as co-winner. Table~\ref{tab:tc_tolerance_sensitivity} reports tolerance-based voting results for $\epsilon = 1.0$. DS-R1-Qwen-7B denotes DeepSeek-R1-Distill-Qwen-7B.
}
\vspace{-2ex}
\label{tab:quant-acc-tc-deltas}
\end{table*}

\clearpage

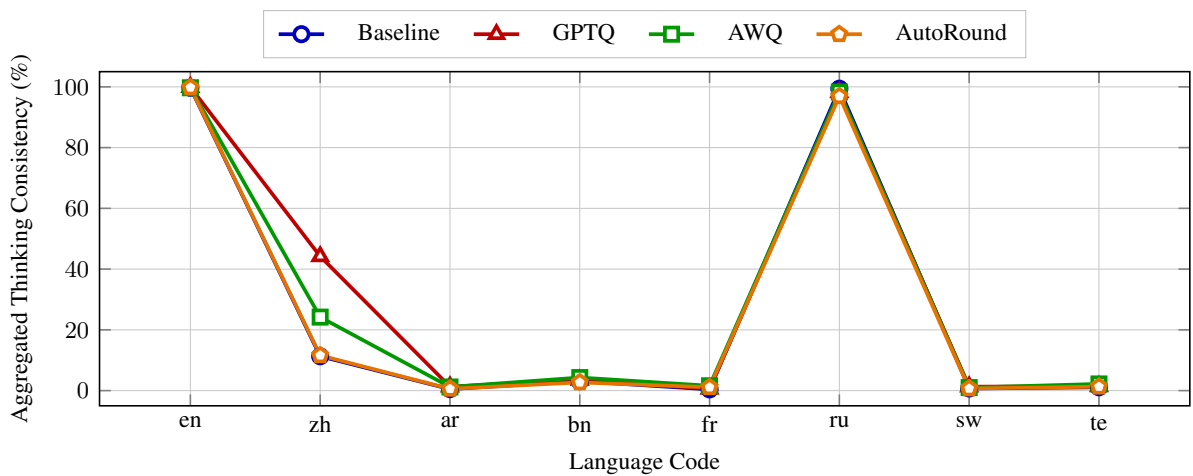
\begin{figure*}[t]
\centering
\begin{tikzpicture}
\begin{axis}[
    width=\textwidth,
    height=6.0cm,
    xlabel={Language Code},
    ylabel={Aggregated Thinking Consistency (\%)},
    ymin=-5, ymax=105,
    xtick=data,
    symbolic x coords={en,zh,ar,bn,fr,ru,sw,te},
    ytick={0,20,40,60,80,100},
    tick label style={font=\small},
    label style={font=\small},
    axis line style={line width=0.6pt},
    tick style={line width=0.6pt},
    grid=major,
    major grid style={
        line width=0.3pt,
        draw=gray!45
    },
    legend style={
        at={(0.5,1.04)},
        anchor=south,
        legend columns=4,
        draw=black!25,
        fill=white,
        font=\small,
        column sep=0.35cm,
        inner xsep=5pt,
        inner ysep=3pt
    },
]

\addplot[
    color=blue!75!black,
    solid,
    mark=*,
    mark options={fill=white, solid},
    line width=1.4pt,
    mark size=2.7pt
] coordinates {
    (en,99.7)
    (zh,11.3)
    (ar,0.5)
    (bn,3.1)
    (fr,0.4)
    (ru,99.4)
    (sw,0.7)
    (te,1.1)
};
\addlegendentry{Baseline}

\addplot[
    color=red!75!black,
    solid,
    mark=triangle*,
    mark options={fill=white, solid},
    line width=1.4pt,
    mark size=3.0pt
] coordinates {
    (en,99.9)
    (zh,44.2)
    (ar,1.3)
    (bn,3.5)
    (fr,0.6)
    (ru,98.2)
    (sw,1.3)
    (te,1.4)
};
\addlegendentry{GPTQ}

\addplot[
    color=green!60!black,
    solid,
    mark=square*,
    mark options={fill=white, solid},
    line width=1.4pt,
    mark size=2.5pt
] coordinates {
    (en,99.7)
    (zh,24.2)
    (ar,1.2)
    (bn,4.3)
    (fr,1.6)
    (ru,98.3)
    (sw,1.0)
    (te,2.2)
};
\addlegendentry{AWQ}

\addplot[
    color=orange!90!black,
    solid,
    mark=pentagon*,
    mark options={fill=white, solid},
    line width=1.4pt,
    mark size=2.8pt
] coordinates {
    (en,99.8)
    (zh,11.6)
    (ar,0.6)
    (bn,2.7)
    (fr,1.1)
    (ru,96.9)
    (sw,0.7)
    (te,1.3)
};
\addlegendentry{AutoRound}

\end{axis}
\end{tikzpicture}

\caption{
Aggregated thinking-language consistency of Baseline and W4A16-quantized
OLMo-3-7B-Think across eight languages.
}
\label{fig:flow-olmo_quantized_tc_eight_languages}
\end{figure*}

\begin{figure*}[t]
\centering
Language Consistency for Chinese with DeepSeek-R1-Distill-Qwen-7B at low difficulty

\vspace{0.15cm}

\includegraphics[width=\linewidth]{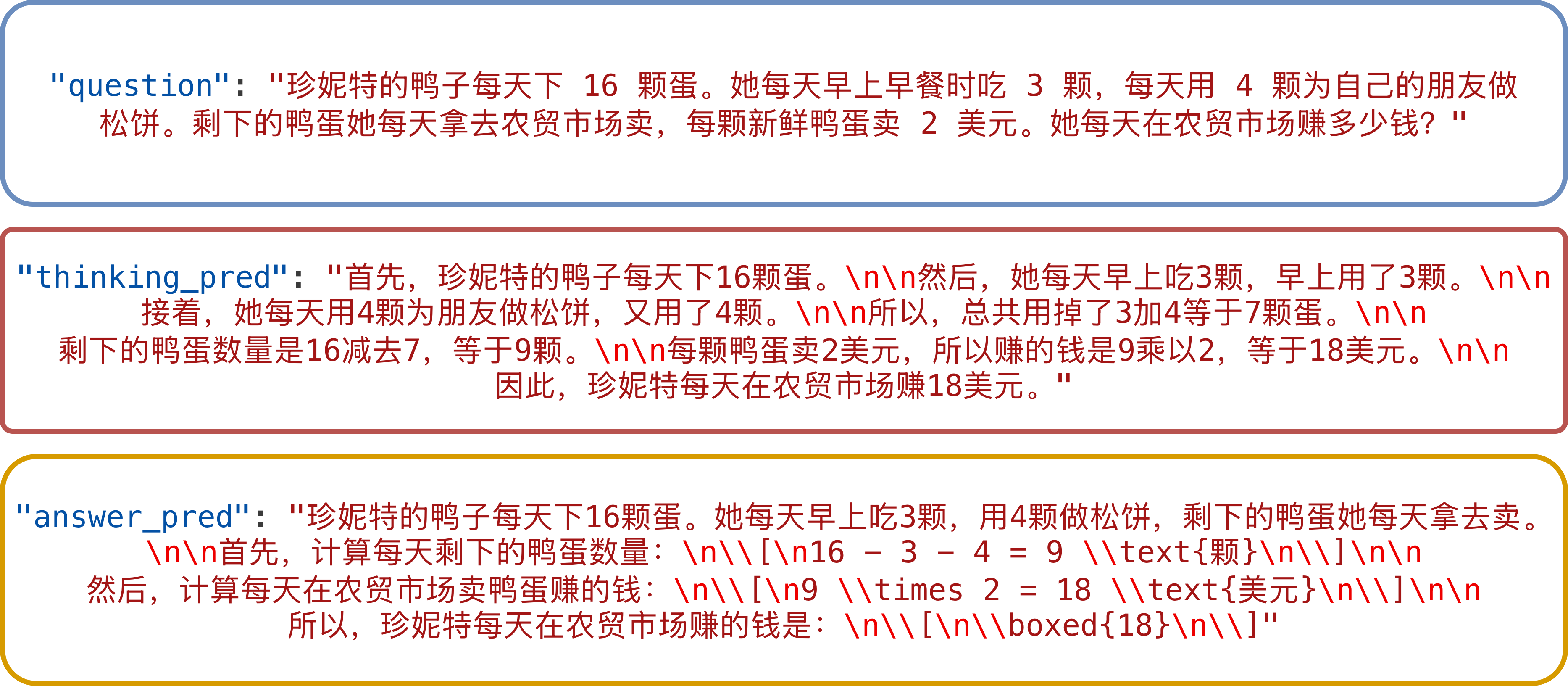}

\caption{Example of output-language consistency: the model receives an Chinese input, reasons and answers in Chinese.}
\label{fig:case1-example-appendix}
\end{figure*}

\begin{figure*}[t]
\centering
Language Inconsistency for Arabic with Qwen3-30B-A3B-Thinking at low difficulty

\vspace{0.15cm}

\includegraphics[width=\linewidth]{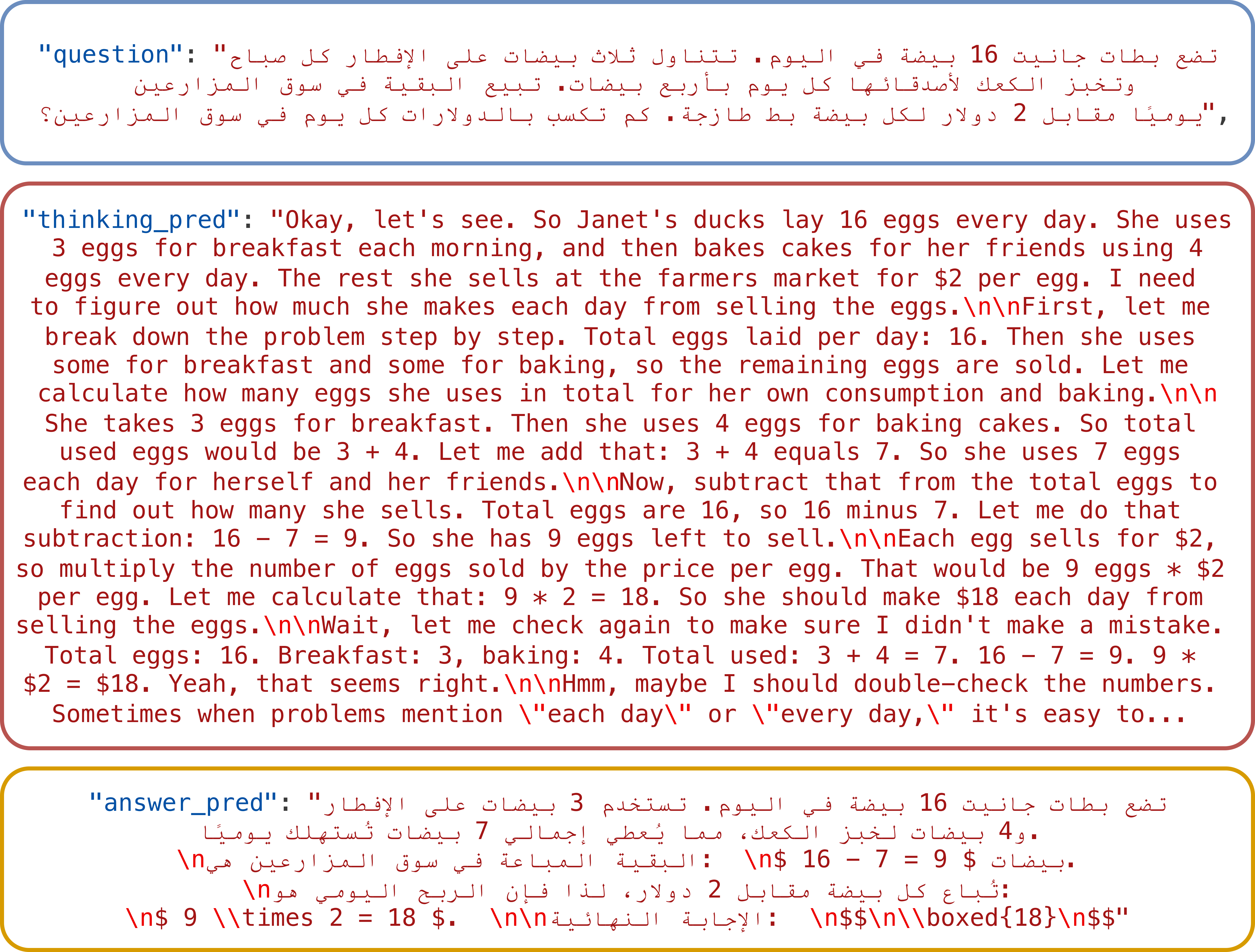}

\caption{Example of output-language inconsistency: the model receives an Arabic input, reasons in English, and answers in Arabic.}
\label{fig:case2-example-appendix}
\end{figure*}

\clearpage

\end{document}